\documentclass[10pt,twocolumn,letterpaper]{article}

\usepackage[pagenumbers]{iccv} % To force page numbers, e.g. for an arXiv version

\usepackage{etoolbox}
\makeatletter
\pretocmd{\chapter}{\addtocontents{toc}{\protect\addvspace{15\p@}}}{}{}
\pretocmd{\section}{\addtocontents{toc}{\protect\addvspace{5\p@}}}{}{}
\pretocmd{\subsection}{\addtocontents{toc}{\protect\addvspace{3\p@}}}{}{}
\makeatother
\usepackage{titletoc}

\usepackage[ruled,algo2e]{algorithm2e}
\usepackage{algorithm}
\usepackage{algorithmic}
\usepackage{comment}
\usepackage{nicefrac} 

\usepackage{amsmath}
\usepackage{amssymb}
\usepackage{mathtools}
\usepackage{amsthm}
\usepackage{comment}
\usepackage{minitoc}

\usepackage{graphicx}
\usepackage{booktabs}
\usepackage{color}

\usepackage{tcolorbox}

\usepackage{cuted}
\usepackage{xpatch}
\usepackage{array}
\usepackage{multirow}
\usepackage{graphicx}
\usepackage{xcolor,colortbl}
\usepackage{pifont}
\usepackage{enumitem}
\usepackage{xcolor}
\PassOptionsToPackage{dvipsnames}{xcolor}
\newcommand{\cmark}{\ding{51}}

\definecolor{Gray}{gray}{0.9}
\newcommand{\highlight}[1]{\cellcolor{Gray}{#1}}

\definecolor{indian red}{RGB}{205,92,92}
\newcommand{\plus}[1]{\small\bf\textcolor{Green}{#1}}
\newcommand{\methodname}{RIG}

\definecolor{iccvblue}{rgb}{0.21,0.49,0.74}
\usepackage[pagebackref,breaklinks,colorlinks,allcolors=iccvblue]{hyperref}

\title{
RIG: Synergizing \underline{R}easoning and \underline{I}magination in End-to-End \underline{G}eneralist Policy
}

\author{
    Zhonghan Zhao$^{1,2*}$\quad Wenwei Zhang$^{2*}$\quad Haian Huang$^{2}$\quad Kuikun Liu$^{2}$ \\\vspace{0pt}
    Jianfei Gao$^{2}$\quad Gaoang Wang$^{1\dag}$\quad  Kai Chen$^{2\dag}$ \vspace{5pt} \\ 
    $^1$Zhejiang University \quad $^2$Shanghai AI Laboratory \\
    \small\texttt{\{zhonghan.22, gaoangwang\}@intl.zju.edu.cn, \{zhangwenwei,chenkai\}@pjlab.org.cn}
}

\begin{document}
\maketitle
\begin{abstract}
Reasoning before action and imagining potential outcomes (i.e., world models) are essential for embodied agents operating in complex open-world environments. Yet, prior work either incorporates only one of these abilities in an end-to-end agent or integrates multiple specialized models into an agent system, limiting the learning efficiency and generalization of the policy. 
Thus, this paper makes the first attempt to synergize \textbf{R}easoning and \textbf{I}magination in an end-to-end \textbf{G}eneralist policy, termed \textbf{\methodname{}}.
To train \methodname{} in an end-to-end manner, we construct a data pipeline that progressively integrates and enriches the content of imagination and reasoning in the trajectories collected from existing agents. The joint learning of reasoning and next image generation explicitly models the inherent correlation between reasoning, action, and dynamics of environments, and thus exhibits more than \textbf{$17\times$} sample efficiency improvements and generalization in comparison with previous works.
During inference, \methodname{} first reasons about the next action, produces potential action, and then predicts the action outcomes, which offers the agent a chance to review and self-correct based on the imagination before taking real actions.
Experimental results show that the synergy of reasoning and imagination not only improves the robustness, generalization, and interoperability of generalist policy but also enables test-time scaling to enhance overall performance.
\end{abstract}
    
\section{Introduction}
\label{sec:intro}

To navigate the complexities of open-world environments,  two quintessential human faculties are \textit{de facto} to embodied agents: imagination of prospective outcomes and reasoning.
Although reasoning endows agents with the ability to deconstruct task objectives into executable plans through logical inference, it inherently operates within the constraints of perceptual history. This limitation underscores the complementarity of world models that learn the environmental dynamics, which not only allows the agent to predict action consequences but also facilitates risk-aware decision-making by evaluating hypothetical trajectories.

\begin{figure}[t]
\centering
    \includegraphics[width=1\linewidth]{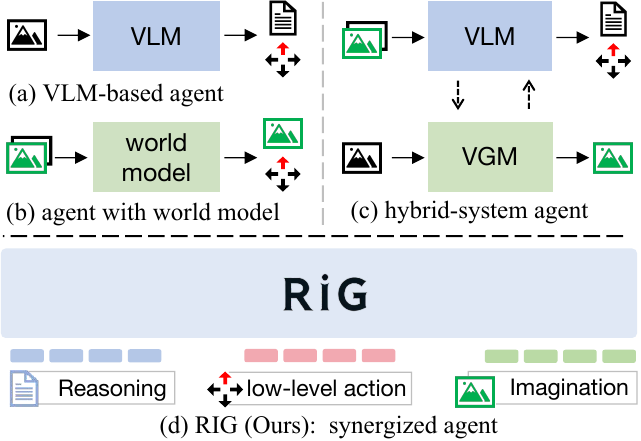}\\
    \vspace{-6pt}
    \caption{\textbf{Comparison between conventional agents and RIG.}
    RIG produces reasoning, actions, and imagination within a single Transformer.}
    \label{fig:introduction}
    \vspace{-18pt}
\end{figure}

The synergistic integration of reasoning and imagination constitutes an indispensable foundation for more intelligent and robust embodied agents operating in dynamically evolving environments.
However, these two abilities are typically implemented in separate models. Specifically, reasoning mainly exists in large vision language models (VLMs)~\cite{zhao2023see,wang2023voyager,wang2023jarvis,zhao2024hierarchical} that parse visual input and produce textual insights and actions (\cref{fig:introduction}(a)), which lack explicit future prediction mechanisms. In contrast, world models~\cite{lin2023learning,hafner2023mastering} specialize in predicting future frames from video data (\cref{fig:introduction}(b)), which suffer from data inefficiency due to the implicit learning of concepts, physical laws, and environment dynamics.
Recent attempts~\cite{zhou2024minedreamer,zhang2023creative,zhao2024we} combine reasoning and imagination by connecting VLMs and visual generative models (VGMs). Yet, the integrated system (\cref{fig:introduction}(c)) prevents end-to-end optimization of the agent, leaving the mutual benefits between reasoning and world models underexplored.

To bridge these gaps, this paper makes the first attempt to synergize \textbf{R}easoning and \textbf{I}magination in an end-to-end \textbf{G}enralist policy, termed \textbf{\methodname{}} ( \cref{fig:introduction}(d)), \methodname{} learns textual reasoning, low-level action control, and image generation through the sequence-to-sequence modeling objective within an autoregressive Transformer, as we hypothesize that the explicit modeling of the logic and motivation behind actions and their consequences could make \methodname{} capture open-world dynamics more comprehensively and improve the sample efficiency of training. 

We develop \methodname{} by adopting a progressive data collection strategy because existing datasets typically lack trajectories that contain interleaved image observations, precise actions, and high-quality textual reasoning. Based on initial trajectories collected from humans~\cite{fan2022minedojo} and existing agents \cite{lifshitz2023steve} that contain only actions and image frames, we first use VLM to insert textual rationales before each action on the trajectory and train \methodname{}-\textit{basic} with the reasoning-enriched trajectories.
During inference, \methodname{}-\textit{basic} generates actions purely from textual and visual inputs, without leveraging imagined future frames, as decisions are executed immediately based on current observations.

To further leverage visual imagination in reasoning to further improve the robustness of the policy, we collect unsuccessful trajectories from \methodname{}-\textit{basic} and adopt GPT-4o to review and revise these trajectories. Then, the suboptimal trajectories are taken as dreamed trajectories and combined with their corresponding revisions to form dream-review style trajectories for training \methodname{} (also noted as \methodname{}-\textit{lookahead} for clarity).
In contrast to \methodname{}-\textit{basic} that conduct reasoning \textit{without} imagination, \methodname{}-\textit{lookahead} learns to first generate a trajectory by taking the predicted images as the environment states, and then review the hypothetical trajectory in reasoning, and predict revised action that changes the environment.
Such a design provides scalability at inference time, where the number of steps in the dream trajectory can be scaled so that the agent can more comprehensively understand the effectiveness of the action and make future-aware decisions.

We extensively evaluate \methodname{} in the diverse, open-world Minecraft environment. Experimental results show that \methodname{} upgrades the state-of-the-art results on embodied tasks, image generation, and reasoning benchmarks by $3.29 \times$,  $2.42 \times$, and $1.33 \times$, respectively. Such a superior performance is achieved by training \methodname{} on only \textbf{111 hours} of videos, which is \textbf{17$\times$} fewer than previous works that rely on $~$2000 hours of videos.
Moreover, when scaling the training data, environmental interactions, and the lookahead steps during reasoning, the generalization ability and robustness of \methodname{} consistently improve,
% The high generalization ability and scalability of \methodname{} 
which implies the potential of synergizing reasoning and imagination in embodied agents.
Our main contributions are summarized as follows: 
\begin{itemize} \item We introduce an end-to-end generalist policy that synergistically integrates explicit reasoning and visual imagination. 
\item We propose a progressive data collection strategy coupled with straightforward language model-based training to efficiently implement our method. 
\item Our method naturally supports test-time scaling, enabling dynamic lookahead reasoning that enhances action robustness and reduces trial-and-error during inference. 
\end{itemize}
\section{Related Work}
\noindent\textbf{Embodied Agents in Minecraft.}
Minecraft presents a significantly open-ended and complex environment~\cite{johnson2016malmo,guss2019minerl,fan2022minedojo,wang2023describe,cai2023open} for embodied agents. Early approaches leveraged explicit world models to predict future states~\cite{hafner2023mastering,cai2023groot} but lack textual reasoning capabilities. Inspired by large language models (LLMs)~\cite{brown2020language,touvron2023llama}, subsequent methods combined LLMs with low-level controllers to address long-horizon tasks. For example, Voyager~\cite{wang2023voyager} and STEVE~\cite{zhao2023see} used LLMs for high-level planning integrated with code databases, while others like Jarvis-1~\cite{wang2023jarvis} paired LLMs with pre-trained low-level policy models such as VPT~\cite{baker2022video}. However, these methods typically lack a world model to explicitly anticipate future visual outcomes. More recently, MineDreamer~\cite{zhou2024minedreamer} integrates a world model and a policy controller, yet treats vision generation and policy control as separate modules, limiting coherent multi-modal reasoning. In contrast, \methodname{} first attempts to explore an end-to-end generalist policy that simultaneously learns textual reasoning, visual imagination, and low-level action predictions to achieve high generalization ability and sample efficiency.

\noindent\textbf{World Models for Embodied Agents.}
Learning robust world models is essential for embodied agents to effectively plan and act within simulated environments~\cite{oh2015actionconditional, kaiser2019model}. Early approaches primarily focused on action-conditioned video prediction or latent imagination for sample-efficient rollouts~\cite{hafner2019dream, hafner2020mastering, schrittwieser2020mastering, hansen2022temporal, lin2023learning}, yet they often tightly coupled the world model with specific policies, limiting their adaptability. Inspired by recent successes in large-scale pre-training~\cite{wu2024pre, mendonca2023structured} and Transformer-based architectures~\cite{micheli2022transformers}, several methods now leverage generalizable knowledge to model visual and textual distributions. However, these models typically overlook explicit reasoning and deeper causal relationships between actions and resulting visual states. \methodname{} explicitly learns to model the joint distribution of textual reasoning, actions, and their visual consequences to enable more accurate predictions of complex and evolving environment dynamics.

\noindent\textbf{Unified Understanding and Generation.}
Multi-modal Large Language models (MLLMs) aim to tackle understanding and generation tasks across different modalities~\cite{lu2023chameleon, zhou2024transfusion} within a unified architecture. Existing methods typically train on large-scale image-text datasets to improve general visual understanding and generation capabilities~\cite{wang2024emu3,yu2023scaling,xie2024show,zhou2024transfusion}. However, these datasets lack the interleaved action and reasoning trajectories required for training embodied agents, limiting their direct applicability to real-world embodied scenarios. Generalist policies like GATO~\cite{reed2022generalist} and RT-1~\cite{brohan2022rt} demonstrate multitask capabilities but optimize each task individually without fully leveraging inter-modal synergies. 
Our work synergizes textual reasoning, low-level action predictions, and visual generation.
\begin{figure*}[t]
\centering
    \includegraphics[width=1\linewidth]{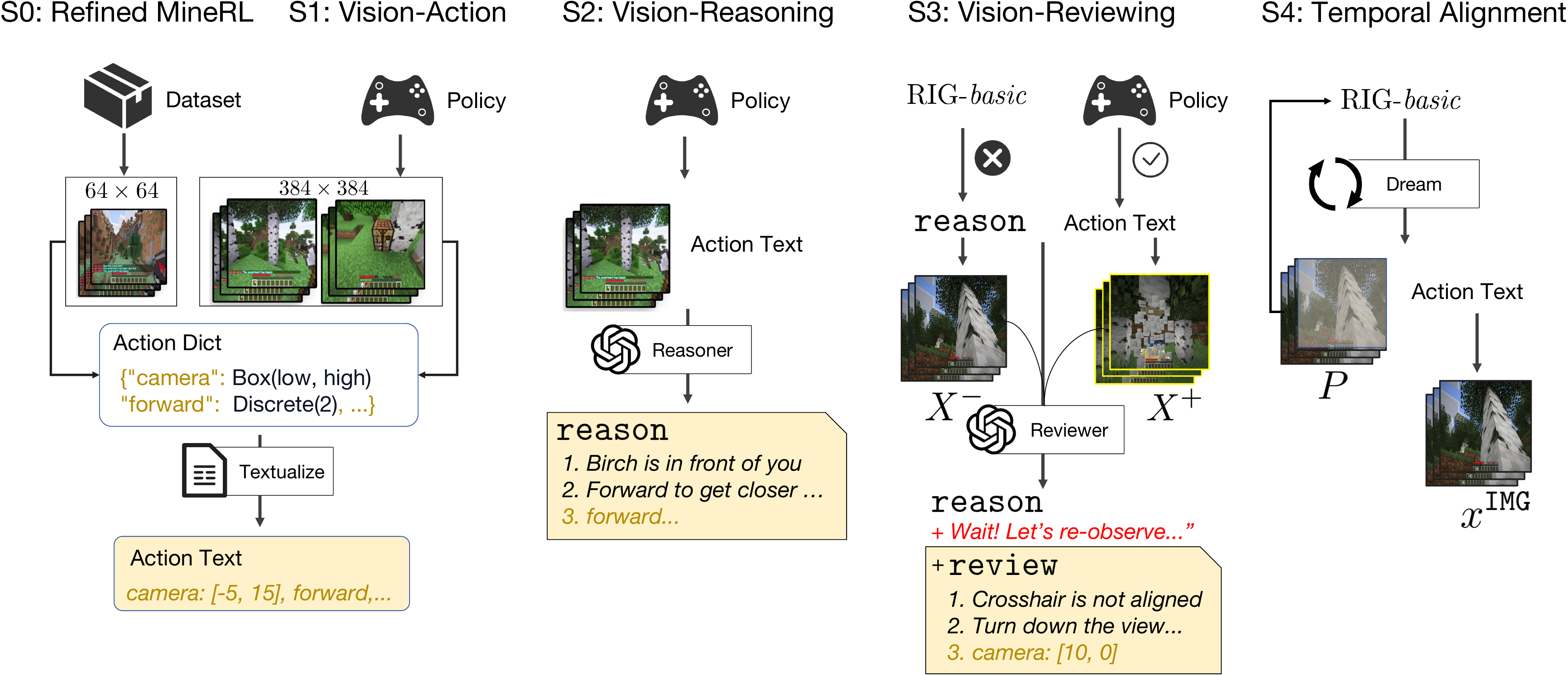}
    \caption{ \textbf{Illustration of the data collection pipeline (S0–S4).} Note that at S3 (Vision-Reviewing), we run the trained \methodname{}-\textit{basic} and policy model (STEVE-1~\cite{lifshitz2023steve}) in parallel, keeping instances where \methodname{}-\textit{basic} performs poorly compared to STEVE-1.}
    \label{fig:data_pipeline}
\end{figure*}

\section{Method}

This paper makes the first attempt to explore the synergy of \textbf{R}easoning and \textbf{I}magination in an end-to-end \textbf{G}eneralist policy, termed  \methodname{}. \methodname{} models image, textual reasoning, and textual action in a sequence-to-sequence manner (\S~\ref{subsec:preliminary}).
We adopt a progressive data collection strategy to first obtain \methodname{}-\textit{basic} that can reason before action but without imagination (\S~\ref{subsec:basic}), then approach \methodname{}-\textit{lookahead} that learns to reason based on generated trajectories (\S~\ref{subsec:review}).

\subsection{Preliminary}\label{subsec:preliminary}
Typical generalist policies follow an autoregressive paradigm to predict actions based on observations~\cite{hafner2023mastering, lin2023learning}. \methodname{} extends this framework by explicitly generating intermediate textual reasoning before action prediction. Specifically, given multi-modal inputs $X = \{ x^{\text{IMG}}, x^{\text{TXT}} \}$ comprising visual tokens $x^{\text{IMG}}$ and textual tokens $x^{\text{TXT}}$, \methodname{} learns to autoregressively generate textual reasoning tokens $Y$, low-level action tokens $A$, and visual prediction tokens $P$:
\begin{equation}
    (Y, A, P) = \mathcal{F}(X),\quad X = \{x^{\text{IMG}}, x^{\text{TXT}}\}.
\end{equation}
The model is trained in an end-to-end manner using only cross-entropy loss:
\begin{equation}
    \mathcal{L} = -\sum_{i=1}\log P_{\theta}(x_i \mid x_{<i}).
\end{equation}
where \( P_{\theta}(\cdot \mid \cdot) \) denotes the conditional probability distribution parameterized by the weights \(\theta\) of \methodname{}.

\subsection{Reasoning without Imagination}\label{subsec:basic}

Our primary goal is to develop a synergized model capable of simultaneously generating textual reasoning, precise low-level actions, and visual outcome predictions. Since existing agents mainly produce actions, existing accessible datasets typically lack comprehensive trajectories containing all these elements. Therefore, we propose a progressive data collection strategy to gradually enrich these elements in accessible agentic trajectories. 
Inspired by the recent success of vision-language models~\cite{gpt4v} that can conduct chain-of-thought~(CoT) reasoning given images, our first step is to add reasoning into the action-image trajectories using VLMs to obtain \methodname{}-\textit{basic} that can conduct reasoning before action.

\noindent\textbf{Data Collection (S0–S2).} As shown in \cref{fig:data_pipeline}, we first refine or collect data from relabeled human play trajectories (S0) and specialized policies (S1), unify their formats, and add reasoning contents before each action (S2). The details are as below:
\begin{itemize}[leftmargin=*]
\item \textbf{S0 (Refined MineRL-V0):}
We use trajectories from MineRL-V0~\cite{guss2019minerl} and quantize the camera actions of the original trajectory into discrete 5-degree intervals and then represent them as textual tokens. All other discrete low-level actions retain their original semantic labels.

\item \textbf{S1 (Vision-Action, 446K):} 
We use a pretrained policy, STEVE-1~\cite{lifshitz2023steve}, to collect high-resolution (384×384) image-action pairs and ensure precise visual-action alignment for learning low-level control.

\item \textbf{S2 (Vision-Reasoning, 200K):} 
To integrate reasoning in the original trajectories, we employ GPT-4o as a \textbf{Reasoner} to annotate explicit textual rationales conditioned on visual observations $x^{\text{IMG}}$ and the corresponding low-level actions $A$, formed as $Y = \textbf{Reasoner}(x^{\text{IMG}}, A)$.
\end{itemize}

\begin{figure}[ht]
\centering
    \includegraphics[width=1\linewidth]{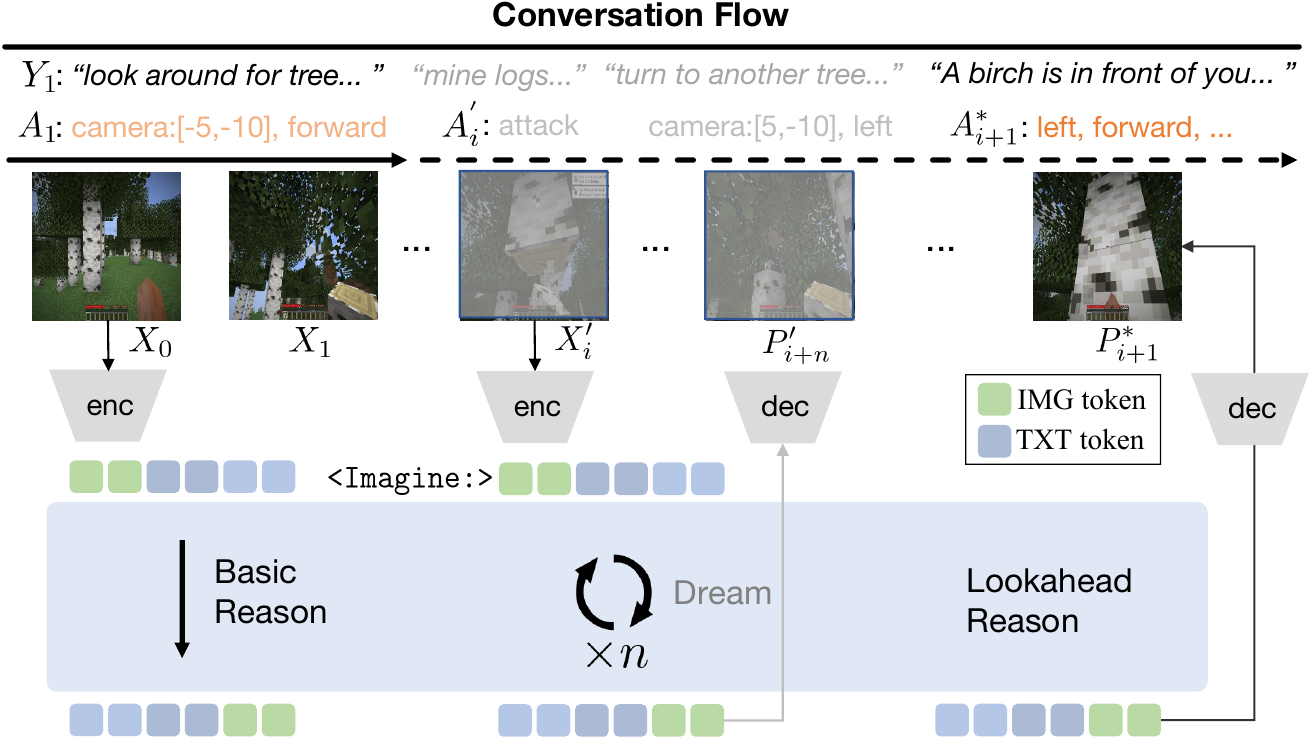}
     \caption{\textbf{Inference process in \methodname{}.} \methodname{} follows a structured \textit{conversation flow} through multi-turn interactions. It consistently uses the fixed word \texttt{Imagine:} to clearly separate internally imagined scenarios from real observations, thereby guiding coherent reasoning, action prediction, and visual imagination.}
    \label{fig:inference}
\end{figure}

All these trajectories are rigorously filtered based on task success, diversity across environment seeds, and manual validation of reasoning quality.
We train \methodname{}-\textit{basic} using datasets obtained from S0, S1, and S2.

\begin{figure*}[t]
\centering
    \includegraphics[width=1\linewidth]{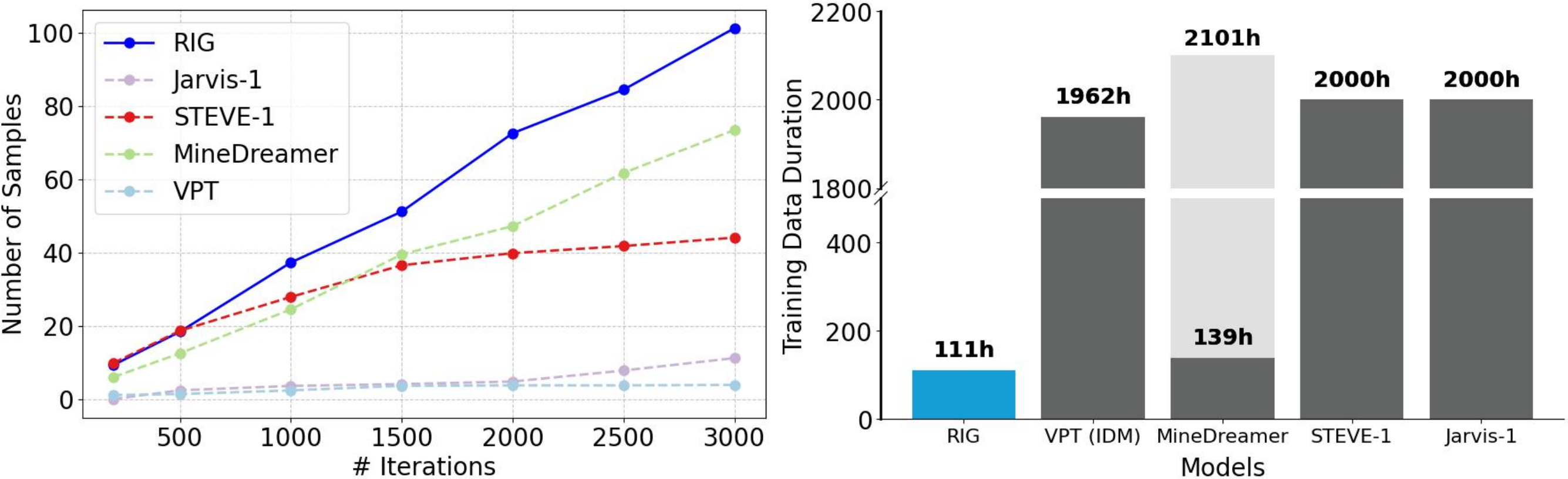}
    \caption{\textbf{Performance and data-efficiency comparison.} \methodname{}-\textit{basic} significantly outperforms other baselines with higher sample efficiency and achieves superior performance using only \textbf{111 hours} of training data (42h S0 MineRL-V0 and 69h S1-S4). MineDreamer~\cite{zhou2024minedreamer}, a hybrid-system model, separately trains a visual generation model (139 hours) but also relies on VPT for the policy model, increasing total data requirements. Duration of VPT~\cite{openai2022vpt} reflects only the IDM data used, measured as video frames, while STEVE-1~\cite{lifshitz2023steve} and Jarvis-1~\cite{wang2023jarvis} also leverage the VPT dataset.}
    \label{fig:data}
\end{figure*}

\noindent\textbf{Reasoning without Imagination.}
After training on datasets (S0, S1, S2), the resulting model, \methodname{}-\textit{basic} naturally supports multi-round interactions with the environment. As shown in \cref{fig:inference}, at each step, it autoregressively generates textual CoT reasoning $Y$, low-level actions $A$, and action outcomes $P$:
\begin{equation}
(Y_{i+1}, A_{i+1}, P_{i+1}) \xleftarrow{\mathcal{F}} (X_i, Y_i, A_i).
\end{equation}
This unified approach achieves significantly better generalization than traditional methods, requiring substantially fewer training samples~(\cref{fig:data}).

\subsection{Lookahead Reasoning}\label{subsec:review}

Although \methodname{}-\textit{basic} demonstrates strong baseline performance, the reasoning is still purely based on the perceptual history and does not fully exploit the generative imagination capabilities. To address this, we further augment our datasets with reflective reviewing annotations in stages 3 and 4 (S3 and S4 in \cref{fig:data_pipeline}), to endow the model with the ability to conduct \textit{lookahead} reasoning, \ie, internally simulate imagined trajectories first, and then take actions after reviewing the predicted future outcomes.

\noindent\textbf{Data Collection (S3–S4).} We collect reflective annotations and temporal alignment data through the following stages:
\begin{itemize}[leftmargin=*]
\item \textbf{S3 (Vision-Reviewing, 27K):}
To fully utilize visual imagination, we introduce a reflective reviewing stage by generating paired trajectories from identical initial states:
\begin{itemize}
\item \textbf{Negative trajectory:} Generated by the previously trained \methodname{}-\textit{basic} model, yielding suboptimal outcomes $X^{-}, Y^{-}, A^{-}$.
\item \textbf{Positive trajectory:} Generated by the superior-performing policy STEVE-1~\cite{lifshitz2023steve}, yielding optimal outcomes $X^{+}, A^{+}$.
\end{itemize}

We then adopt GPT-4o as a \textbf{Reviewer} to explicitly compare these parallel trajectories and generate refined reasoning: $Y^{+} = \textbf{Reviewer}(X^{-}, Y^{-}, A^{-}, A^{+})$, so that we get corrective reasoning annotations:
\begin{equation}
    Y = \{Y^{-}, \text{``Wait! Let's re-observe...''}, Y^{+}\}.
\end{equation}
This reflection annotation significantly enhances the ability of the model to review and correct reasoning mistakes.

\item \textbf{S4 (Temporal Alignment, 38K):}
We further generate multi-step imagined visual predictions ($P$) and explicitly align them with observed ground-truth visual tokens ($x^{\text{IMG}}$) to enhance long-horizon stability: $P_{i+1} \rightarrow x^{\text{IMG}}_{i+1}.$

\end{itemize}

\noindent\textbf{Lookahead Reasoning with Imagination.}
Training on datasets from stages 3 and 4 produces \methodname{}-\textit{lookahead}, a model that performs reasoning conditioned on imagined futures. stage 3 adopts Rejection Sampling Fine-tuning (RFT) to improve reasoning through model-generated rollouts. We apply RFT in embodied agents by leveraging joint reasoning and visual generation, which enables self-prediction of future states, previously infeasible due to the lack of visual prediction. Only the positive trajectory $Y^{+}$ is optimized, while the negative $Y^{-}$ is excluded from loss, encouraging better self-correction.

\methodname{}-\textit{lookahead} simulates ``dream trajectories'' before acting. As shown in \cref{fig:inference}, imagined steps are marked with a fixed token ``\texttt{<Imagine:>}'' to distinguish from observations, allowing decisions to be refined by looking $n$ steps ahead:

\begin{equation}
(Y^{*}_{i+1}, A^{*}_{i+1}, P^{*}_{i+1}) \xleftarrow{\mathcal{F}} (X_i, P_{i+1}, Y_{i+1}, ..., P_{i+n}, Y_{i+n}).
\end{equation}

This lookahead mechanism enables internal review and correction, reducing trial-and-error interactions and enhancing decision robustness in complex embodied tasks, as shown in~\cref{fig:example}.

\section{Experiments}
We conduct comprehensive experiments to validate the effectiveness of \methodname{} across diverse tasks, focusing on data efficiency, scalability, and the benefits of integrating generation, reasoning, and lookahead. Evaluations on embodied tasks are performed under both \textit{Manual} (hand-only) and \textit{Tool} (\eg, iron pickaxe) to assess performance in varied embodied scenarios.

\subsection{Inplementation Details}

\methodname{} is initialized from the pretrained Janus-1.4B~\cite{chen2025janus} with a sequence length of 4096 tokens. It operates as a fully end-to-end agent, integrating textual reasoning, visual understanding, planning, and self-review within a unified Transformer model. For visual understanding, we use SigLIP-Large-Patch16-384~\cite{zhai2023sigmoid}, while for visual generation RIG employs a VQ-based encoder with a 16,384-codebook and a 16$\times$ downsampling factor, each with two-layer MLP adaptors. A VQ tokenizer~\cite{sun2024autoregressive} converts images into discrete IDs, which are embedded and concatenated with text for multi-modal processing. The training utilizes sequence packing and mixed data types, conducted on XTuner-lite~\cite{2023xtuner}. 
The experiments on embodied tasks mainly follow the setup of STEVE-1~\cite{lifshitz2023steve}, evaluated by the number of samples for collection tasks and accuracy for exploration, both derived from environment feedback.

\begin{figure*}[t]
\centering
    \includegraphics[width=1\linewidth]{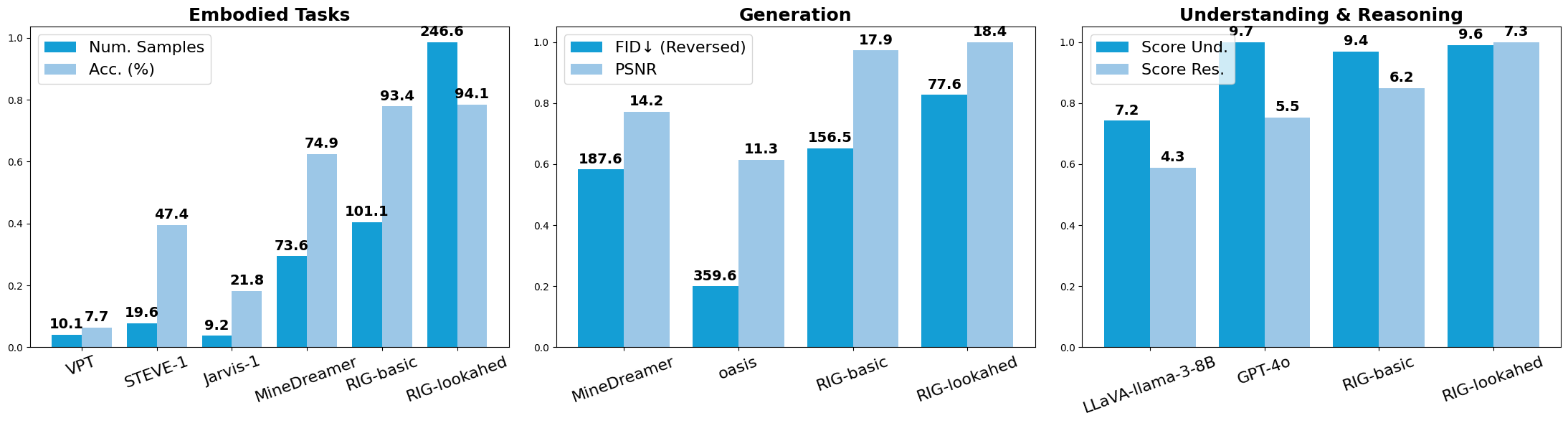}\\
    \caption{\textbf{Comparison with various baselines across embodied tasks, generation, understanding, and reasoning.} \methodname{}-\textit{basic} incorporates reasoning without reviewing, while \methodname{}-\textit{lookahead} integrates both reasoning and reviewing capabilities.}
    \label{fig:main}
\end{figure*}

\begin{figure*}[t]
\centering
    \includegraphics[width=1\linewidth]{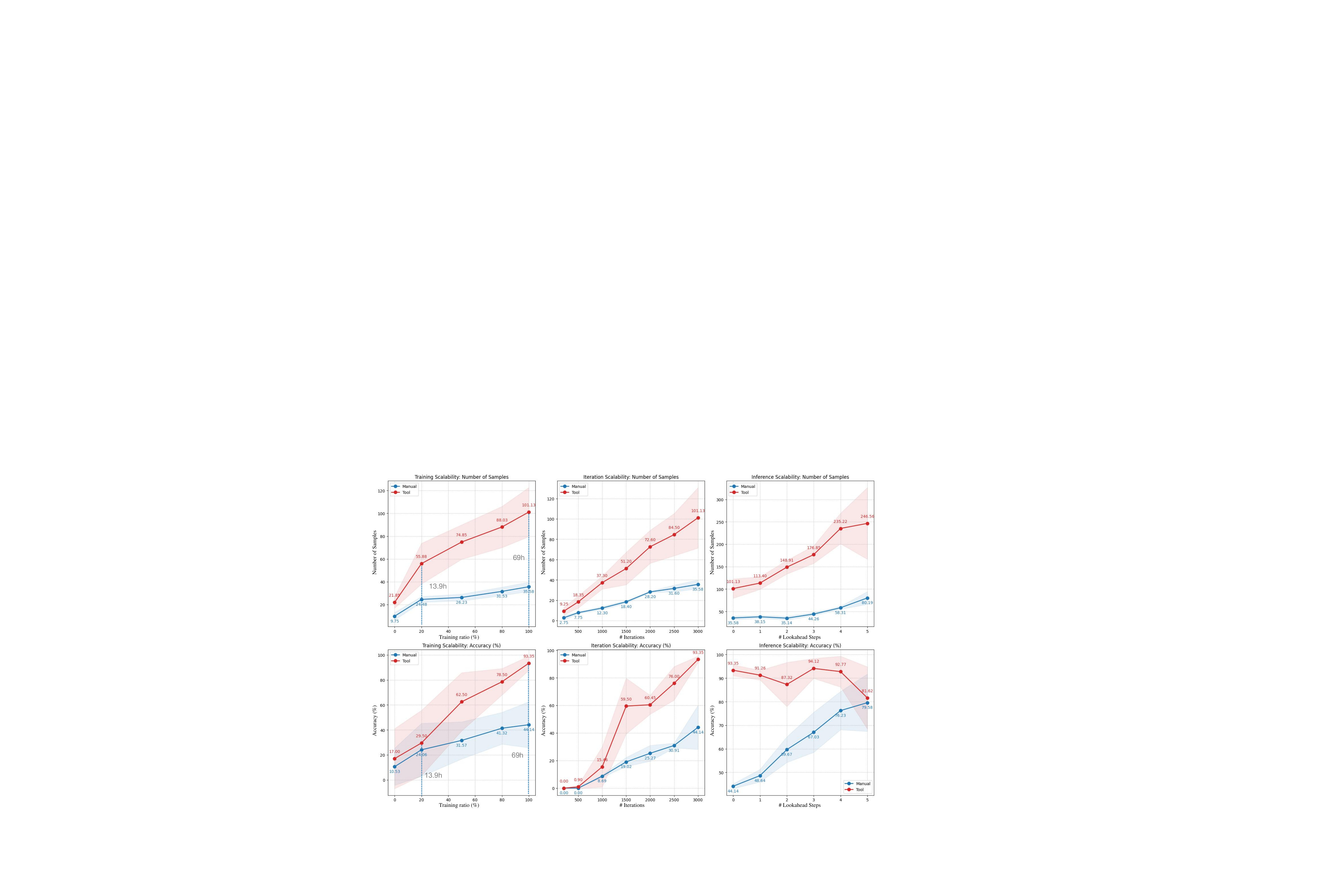}\\
   \caption{\textbf{Scalability Evaluation Across Training, Iteration, and Inference.} 
    We evaluate the scalability of \methodname{} by testing its performance across three different aspects: \textbf{training scalability}, \textbf{iteration scalability}, and \textbf{inference scalability}. 
    Each column corresponds to a different scalability setting. The top row presents the number of collected samples in material-gathering tasks, while the bottom row reports the success rate in exploration-based tasks. 
    Shaded regions represent variance. We exclude 42h MineRL-V0 pretraining from the total 111h in Figure. The training ratio is only counted before the lookahead reasoning.}
    \label{fig:scale}
\end{figure*}

\begin{table*}[t!]
\centering
\resizebox{\textwidth}{!}{
\begin{tabular}{l|cccc|ccccc|ccccc}
\toprule
\multirow{2}{*}{ID} 
& \multicolumn{4}{c|}{Capabilities} 
& \multicolumn{5}{c|}{Number of Samples} 
& \multicolumn{5}{c}{Accuracy (\%)} \\
\cmidrule(lr){2-5} \cmidrule(lr){6-10} \cmidrule(lr){11-15}
& Action & Gen. & Reason & Lookahead 
& wood & grass & dirt & avg. & $\Delta$ 
& Dig & Explore & Tower & avg. & $\Delta$ \\
\midrule
\rowcolor{Gray}
\multicolumn{15}{l}{\textit{Manual (ID 0--4)}} \\
\midrule
0 & \cmark &       &       &       & 7.9  & 6.2  & 8.9  & 7.7  & \plus{+0.0} & 9.1  & 11.7 & 4.4  & 8.4  & \plus{+0.0} \\
1 & \cmark & \cmark &       &       & 11.0 & 16.5 & 12.1 & 13.2 & \plus{+5.5}  & 12.2 & 36.8 & 41.8 & 30.3 & \plus{+21.9} \\
2 & \cmark &       & \cmark &       & 17.3 & 24.5 & 22.5 & 21.4 & \plus{+13.8} & 34.2 & 31.8 & 37.8 & 34.6 & \plus{+26.2} \\
3 & \cmark & \cmark & \cmark &       & 22.2 & 45.9 & 38.7 & 35.6 & \plus{+27.9} & \textbf{29.2} & 65.2 & 37.9 & 44.1 & \plus{+35.7} \\
4 & \cmark & \cmark & \cmark & \cmark
  & \textbf{28.3} & \textbf{137.5} & \textbf{74.8} & \textbf{80.2} & \plus{+72.5}
  & 65.8 & \textbf{84.2} & \textbf{88.7} & \textbf{79.6} & \plus{+71.2} \\
\midrule
\rowcolor{Gray}
\multicolumn{15}{l}{\textit{Tool (ID 0--4)}} \\
\midrule
0 & \cmark &       &       &       & 24.6 & 33.1 & 42.4 & 33.4 & \plus{+0.0}  & 17.9 & 11.7 & 8.2  & 12.6 & \plus{+0.0} \\
1 & \cmark & \cmark &       &       & 25.8 & 29.9 & 48.4 & 34.7 & \plus{+1.3}  & 27.2 & 36.8 & 41.8 & 35.3 & \plus{+22.7} \\
2 & \cmark &       & \cmark &       & 26.9 & 49.9 & 51.0 & 42.6 & \plus{+9.2}  & 24.2 & 31.8 & 29.8 & 28.6 & \plus{+16.0} \\
3 & \cmark & \cmark & \cmark &       & 79.4 & 115.3 & 108.6 & 101.1 & \plus{+67.7} & 85.1 & 100.4 & 94.7 & 93.4 & \plus{+80.8} \\
4 & \cmark & \cmark & \cmark & \cmark
  & \textbf{128.7} & \textbf{295.6} & \textbf{315.5} & \textbf{246.6} & \plus{+213.2}
  & \textbf{95.4} & \textbf{84.2} & \textbf{102.8} & \textbf{94.1} & \plus{+81.5 } \\

\bottomrule
\end{tabular}
}
\caption{\textbf{Ablation study on embodied tasks under different capability settings.} We compare different combinations of Action, Generation (Gen.), Reasoning (Reason), and Reviewing (Review). The table is divided into two groups: \emph{Manual} (ID 0–4) and \emph{Tool} (ID 0-4). The ``Num.'' column represents the number of completed collecting tasks (wood, grass, dirt), while ``Acc.'' denotes the success rate of exploration tasks. The columns ``avg.'' is the average performance. For both metrics, we report the absolute values, along with the improvement (\plus{+}$x$) over the baseline (ID 0 for \emph{Manual} and \emph{Tool}).
}
\label{tab:ablation_embodied}
\end{table*}

\begin{table*}[t!]
    \centering
    \resizebox{\linewidth}{!}{
    \begin{tabular}{
        l|cccc|
        >{\centering\arraybackslash}p{1.6cm}
        >{\centering\arraybackslash}p{1.6cm}|
        >{\centering\arraybackslash}p{1.8cm}
        >{\centering\arraybackslash}p{1.8cm}|
        >{\centering\arraybackslash}p{1.8cm}
    }
    \toprule
    \multirow{2}{*}{ID} & 
    \multicolumn{4}{c|}{Capabilities} & 
    \multicolumn{2}{c|}{Generation} &
    \multicolumn{2}{c|}{Understanding} &
    Reasoning \\
    \cmidrule{2-10}
    & Action & Gen. & Reason & Lookahead
    & FID $\downarrow$ & PSNR $\uparrow$
    & Score-Stc.$\uparrow$ & Score-Env.$\uparrow$ & Score-Env.$\uparrow$ \\
    \midrule
    0 &  & \cmark &  &  & \highlight{214.5}     & \highlight{16.4}     & -   & -   & -    \\
    1 & \cmark & \cmark &  &  & \highlight{225.6} & \highlight{16.3} & -   & -   & -    \\
    \midrule    
    2 & \cmark &  & \cmark &   & -     & -     & \highlight{9.0} & \highlight{7.8} & \highlight{6.1}  \\
    3 & \cmark & \cmark & \cmark &  & 156.5 & 17.9 & \highlight{9.4} & \highlight{\textbf{8.4}} & \highlight{7.3}  \\
    \midrule
    4 & \cmark & \cmark & \cmark & \cmark  & \textbf{77.6}  & \textbf{18.4} & \textbf{9.6} & 8.1 & \textbf{8.5} \\
    \bottomrule
    \end{tabular}
    }
    \caption{\textbf{Ablation study on Generation, Understanding, and Reasoning performance.} We compare different combinations of Action, Generation (Gen.), Reasoning (Reason), and Lookahead capabilities. ``Score-Env.'' represents the environment-specific evaluation score from online understanding testing, while ``Score-Env.'' denotes reasoning-specific evaluation. ``Score-Stc.'' is computed on the static dataset STEVE-21K~\cite{zhao2023see}, and ``FID'' / ``PSNR'' measure image generation quality.}
    \label{tab:ablation_others}
\end{table*}

\subsection{Main Results}
We first evaluate the data efficiency of \methodname{} against prior approaches (VPT~\cite{openai2022vpt}, STEVE-1~\cite{lifshitz2023steve}, Jarvis-1~\cite{wang2023jarvis}, and MineDreamer~\cite{zhou2024minedreamer}). We then benchmark its performance across three core task categories: Embodied Tasks, Generation Tasks, and Understanding \& Reasoning Tasks. For more qualitative analysis, please refer to the case study in~\cref{fig:case}.

\noindent\textbf{Data Efficiency and Training Duration.}
As shown in \cref{fig:data}, \methodname{} significantly surpasses other methods in terms of collected samples per iteration while requiring drastically lower total training time. Notably, \methodname{} achieves superior performance using only \textbf{111 hours} of total data collection, considerably less than other baselines (VPT: 1962h, MineDreamer: 2101h (139h + VPT), STEVE-1 and Jarvis-1: nearly 2000h).

\noindent\textbf{Performance in Embodied Tasks.} As shown in \cref{fig:data}, \methodname{}-\textit{basic} have surpassed all other baselines with \textbf{93.4\%} accuracy and \textbf{101.1} collected samples. \methodname{}-\textit{lookahead} (extended from \methodname{}-\textit{basic}) produce even greater progress than \methodname{}-\textit{basic}, with the highest accuracy of \textbf{94.1\%} and \textbf{246.6 collected samples}. For extended analysis, we conduct additional scalability and ablation studies, as shown in \cref{fig:scale} and \cref{tab:ablation_embodied}.

\noindent\textbf{Performance in Generation Tasks.}
As shown in~\cref{fig:main}, \methodname{}-\textit{lookahead} achieves the best generation quality among all baselines, with the lowest FID (\textbf{77.6}) and highest PSNR (\textbf{18.4}) compared to MineDreamer and Oasis. As shown in~\cref{tab:ablation_others}, performance consistently improves as capabilities are progressively integrated, with the most substantial gains attributed to the inclusion of lookahead reasoning. \methodname{}-\textit{lookahead} significantly enhances generation performance over variants without it (\eg, ID3, FID: 156.5, PSNR: 17.9).

\noindent\textbf{Performance in Understanding \& Reasoning.}
As shown in~\cref{fig:main}, \methodname{}-\textit{lookahead} demonstrates superior reasoning capabilities, achieving a Reasoning score of \textbf{7.3}, surpassing GPT-4o (5.5), and an Understanding score of \textbf{9.6}, on par with GPT-4o (9.7). Additionally,~\cref{tab:ablation_others} further validates the benefit of synergizing components, as reasoning and reviewing capabilities jointly contribute to stronger task understanding.

\subsection{Analysis of Scalability}~\label{sec:scale}
We evaluate the scalability of \methodname{} along three key dimensions, training data ratio, iteration count, and inference steps, as illustrated in \cref{fig:scale}.

\noindent\textbf{Training Scalability.} \methodname{} exhibits strong scalability with training data volume. Increasing training data from 10\% to 100\% dramatically improves performance, especially at the 20\% training threshold, where accuracy jumps substantially (manual: 10.53\%→24.06\%, tool: 17.0\%→62.50\%). Data diversity significantly increases at this point, allowing the agent to encounter and adapt to a broader spectrum of complex scenarios. Beyond 20\% data usage, the rate of accuracy improvement stabilizes, indicating the training paradigm reaches a steady state and that the agent's action diversity nears its upper bound. With full training data (100\%), \methodname{} achieves superior results, outperforming existing approaches such as VPT~\cite{openai2022vpt} and Jarvis-1~\cite{wang2023jarvis} in accuracy and even surpassing STEVE-1~\cite{lifshitz2023steve} in task collection efficiency, reaching 101.13 collected tasks and 93.35\% accuracy. Notably, these results are attained purely through forward reasoning, without lookahead, suggesting substantial untapped potential for further enhancement by incorporating advanced reasoning mechanisms.

\noindent\textbf{Iteration Scalability.} \methodname{} demonstrates robust performance growth over iterations. Under standard forward inference, task collection grows consistently from 9.25 samples at iteration 200 to 101.13 at iteration 3000, particularly pronounced in tool-assisted tasks, showing rapid convergence due to effective data utilization and stable trajectory patterns. However, variance, illustrated by shaded areas, tends to increase with iterations, reflecting longer and more diverse trajectories that introduce complexity and fluctuation. Tasks involving exploration, which inherently contain more combinatorial subtasks (\eg, material gathering followed by building structures), show larger variance and complexity over time. It potentially highlights the model’s adaptive response to increasingly diverse scenarios.

\noindent\textbf{Inference Scalability.} The results from different lookahead steps demonstrate significant benefits from lookahead reasoning. Evaluating from the baseline at 3000 iterations, increasing steps (generating ``dream trajectories'') substantially improves performance. Task collection metrics exhibit rapid initial improvement and relatively low variance up to four steps, indicating accurate and stable trajectory predictions. However, variance increases at five steps, suggesting accumulated prediction errors or hallucinations become more prominent. For accuracy metrics, tool-assisted tasks maintain high performance (peaking at 94.12\% at 3 steps), with a slight decrease afterward due to ceiling effects and increased prediction uncertainty. Conversely, manual tasks show consistent performance improvement through stepwise lookahead, significantly benefiting from iterative reasoning, reaching a peak of 79.58\% accuracy at 5 steps.

\subsection{Ablation Study on Embodied Tasks}

\noindent\textbf{Effect of Generation.}
Incorporating visual generation significantly boosts performance, especially in exploration-based tasks. Comparing ID 0 and ID 1, the addition of visual generation yields clear gains in resource collection (\textit{Number of Samples}, +5.5) and exploration accuracy (\textit{Acc.}, +21.9) under the Manual setting, with corresponding improvements of +1.3 and +22.7 under the Tool setting. Visual generation improves the model’s ability to interpret action outcomes, particularly in tasks requiring precise targeting. For example, misalignment of the crosshair in mining tasks can drastically affect success. Generating future visual states enables better action-state alignment, leading to more informed decisions.

\noindent\textbf{Effect of Reasoning.}
Incorporating explicit reasoning yields substantial improvements across both Manual and Tool settings. Compared to ID 0, ID 3 with reasoning increases the \textit{Number of Samples} from 7.7 to 35.6 (+27.9) and accuracy by +35.7 in the Manual setting; in the Tool setting, it boosts sample count by +67.7 and accuracy by +80.8. These gains arise from two key effects: (1) reducing redundant actions, without reasoning, autoregressive policies tend to overuse actions like ``attack’’; and (2) enabling more strategic, goal-directed exploration, reasoning helps the model better interpret its environment, especially in multi-step tasks like \textit{Tower}, where resource gathering and construction must be planned jointly. When combined with generation (ID 2), even earlier variants already show consistent performance gains (\eg, Manual: +13.8 samples, +21.9 accuracy), highlighting the synergistic value of reasoning.

\noindent\textbf{Effect of Lookahead Reasoning.}
Introducing reviewing brings the most significant gains. In the Manual setting, ID 4 achieves the best performance: \textit{Number of Samples} rises from 7.7 to 80.2 (+72.5), and accuracy reaches 79.6\% (+71.2). In the Tool setting, improvements are even more pronounced: from 33.4 to 246.6 (+213.2) in sample collection and from 12.6\% to 94.1\% (+81.5) in accuracy. As illustrated in~\cref{fig:inference}, self-reviewing enables the agent to anticipate and reason over imagined future states, refining actions before execution. Notably, this capability only requires a small amount of additional data, 27K samples (0.8 hours), compared to thousands of hours used by other methods, demonstrating high efficiency.

\noindent\textbf{Effect of Synergizing.}
As shown in~\cref{tab:ablation_embodied}, the synergy of \textit{Action}, \textit{Generation}, \textit{Basic Reasoning} and \textit{Lookahead Reasoning} leads to the most robust performance, which enables structured learning and improves short-term decisions and long-horizon task completion. In the Manual setting (ID 4), there are significant gains in sample collection (from 7.7 to 80.2, +72.5) and accuracy (from 8.4\% to 79.6\%, +71.2). In the Tool setting, the impact is even greater, with sample counts rising from 33.4 to 246.6 (+213.2) and accuracy from 12.6\% to 94.1\% (+81.5), highlighting the effectiveness of the unified framework.

\subsection{Ablation Study on Generation Quality}

\noindent\textbf{Effect of Generation.}
As shown in \cref{tab:ablation_others}, visual generation significantly impacts overall system performance. 
Comparing ID 0 and ID 1, introducing action capability slightly degrades FID performance (from 214.5 to 225.6). This suggests that simply incorporating action learning without proper synergy with reasoning may introduce inconsistencies in the learned visual representations. However, when both action and reasoning are enabled (ID 3 and ID 4), generation quality improves substantially, with FID dropping to 156.5 and PSNR rising to 17.9.

\noindent\textbf{Effect of Reasoning.}
Adding reasoning capabilities (ID 2 and ID 3) improves understanding and environmental interaction. Comparing ID 2 with ID 3, we see an increase in \textit{Score-Static} (9.0 to 9.4) and \textit{Score-Env.} (6.1 to 7.3). This demonstrates that reasoning enhances decision-making and allows for more informed perception of environmental dynamics. When reasoning is combined with generation (ID 3), \methodname{} benefits from enhanced action planning and prediction, further improving overall performance.

\noindent\textbf{Effect of Lookahead Reasoning.}
Including lookahead reasoning (ID 4) results in the best performance across all evaluation metrics. FID improves significantly (from 156.5 to 77.6), PSNR reaches 18.4, and \textit{Score-Static} and \textit{Score-Env.} achieve their highest values at 9.6 and 8.5, respectively. Lookahead reasoning enhances visual prediction by improving reasoning about future visual states.

\noindent\textbf{Effect of Synergizing Reasoning and Imagination.}
By jointly optimizing generation, reasoning, and reviewing, \methodname{} achieves the best trade-off between action prediction, visual understanding, and environmental reasoning. The best model (ID 4) shows that lookahead reasoning enhances decision-making, improving sample efficiency and interaction robustness. This underscores the benefit of integrating multiple modalities for coherent perception and action. As shown in \cref{tab:ablation_others}, enabling lookahead reasoning substantially enhances image-generation quality. \methodname{}-\textit{lookahead} (ID 4) attains the lowest FID (77.6) and highest PSNR (18.4), significantly surpassing variants without lookahead reasoning (\eg, ID 3, FID: 156.5, PSNR: 17.9).
\section{Conclusion}

This paper introduces \methodname{}, an end-to-end \textbf{G}eneralist policy that integrates \textbf{R}easoning and \textbf{I}magination with superior adaptability and robustness in open-world environments. \methodname{} unifies the understanding and generation of visual generation, action, and textual reasoning within a single autoregressive Transformer, and is capable of reasoning and planning by looking ahead with the dreamed trajectories to further improve its robustness.
\methodname{} obtains new state-of-the-art performance across embodied tasks, image generation, and reasoning tasks, with higher sample efficiency and generalization. \methodname{} also exhibits higher scalability with training and test-time compute.
We hope the results of \methodname{} could inspire future research on synergizing reasoning and imagination in embodied agents.

\appendix
\renewcommand\thefigure{\Alph{section}\arabic{figure}}
\renewcommand\thetable{\Alph{section}\arabic{table}}
\setcounter{figure}{0}
\setcounter{table}{0}

\renewcommand{\thesection}{A}

\section{Appendix}

\paragraph{The appendix is organized as follows:}
\begin{itemize}

\item[$\bullet$] \textbf{Inference Pipeline}~(\cref{sec:inferencedetails}) illustrates how \methodname{} performs end-to-end inference: generating textual reasoning, imagined visual rollouts, and executable actions, with explicit use of the \texttt{<Imagine:>} token to support self-review and temporal consistency.

\item[$\bullet$] \textbf{Training Pipeline}~(\cref{sec:training}) presents our multi-stage training framework, including offline supervised learning, GPT-4o-based reasoning and review relabeling, and imagination-grounded alignment strategies that enable lookahead-based decision-making.

\item[$\bullet$] \textbf{Data Distribution}~(\cref{sec:data_distribution}) analyzes the diversity of embodied tasks within our datasets, and illustrates how data volume scales with task complexity—from atomic skills like collection to composite ones like exploration and construction.

\item[$\bullet$] \textbf{Component Comparison}~(\cref{sec:component}) offers a systematic comparison with existing models, emphasizing \methodname{}’s unique capabilities in multimodal alignment, action granularity, and unified policy formulation without relying on task-specific modules.

\item[$\bullet$] \textbf{Tokenizer and Base Model Selection}~(\cref{sec:tokenizer}) explains our design choice of combining LlamaGen’s VQ tokenizer with Janus as the vision-language foundation, offering a lightweight and effective setup for image-text grounding in Minecraft-like settings.

\item[$\bullet$] \textbf{Qualitative Results and Case Study}~(\cref{sec:demo}) showcases examples where \methodname{} performs internal reasoning, detects failure cases via self-review, and corrects actions before execution. We further compare it to GPT-4o, demonstrating that strong visual generation alone does not guarantee robust policy reasoning.

\item[$\bullet$] \textbf{Multi-Modal Understanding Evaluation}~(\cref{sec:under_eval}) evaluates \methodname{}’s embodied knowledge across diverse functional categories using the STEVE-21K QA benchmark, covering survival, crafting, entity understanding, and more.

\item[$\bullet$] \textbf{Multi-turn Visual Reasoning Format}~(\cref{sec:prompt_format}) details our multi-round reasoning and imagination format, which supports fine-grained learning of vision-language-action alignment through step-by-step trajectory prediction.

\item[$\bullet$] \textbf{Environment Details}~(\cref{sec:env_detail}) describes our experimental platform based on MineRL~\cite{guss2019minerl}, featuring low-level egocentric control and programmable environment setup for robust and reproducible embodied evaluation.

\end{itemize}

\begin{figure}[h]
\centering
\includegraphics[width=1\linewidth]{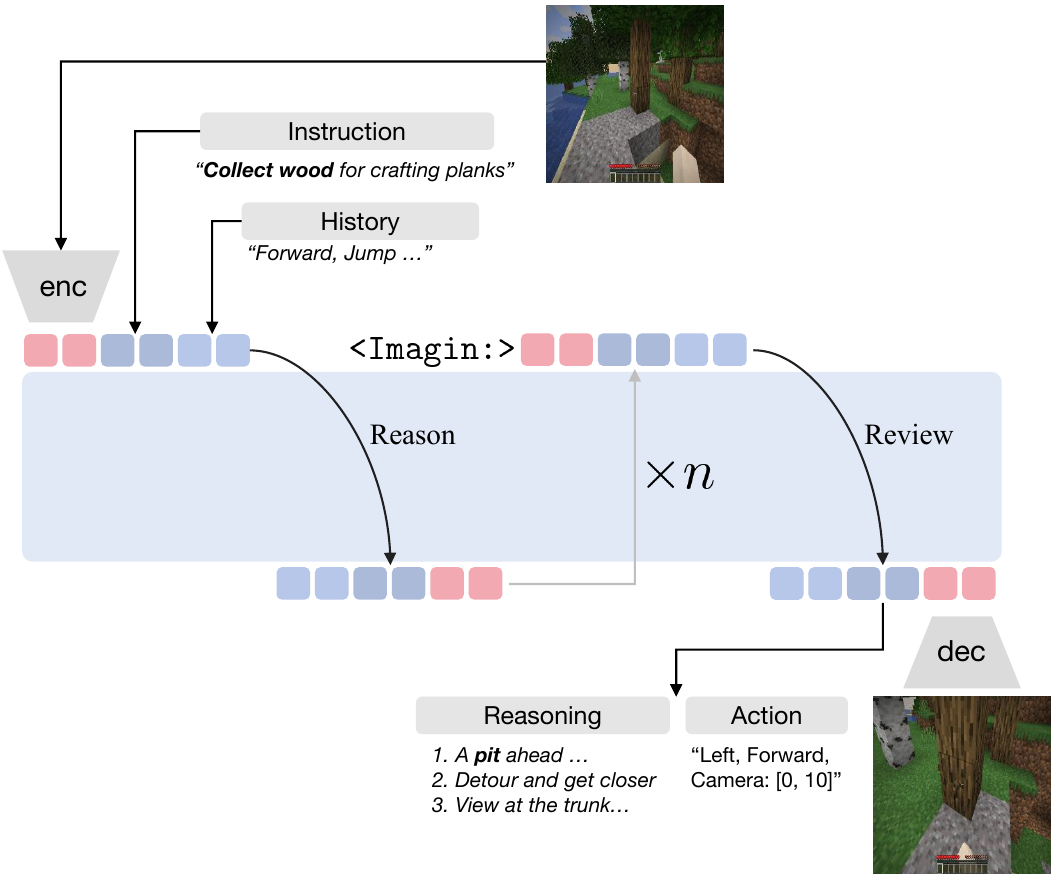}
\caption{\textbf{Detailed inference pipeline.} \methodname{} generates imagined visual states and corresponding reasoning to simulate multiple action trajectories, enabling self-review and corrective prediction.}
\label{fig:inference_d}
\end{figure}

\subsection{Inference Pipeline}\label{sec:inferencedetails}
As illustrated in \cref{fig:inference_d}, \methodname{} follows a multimodal autoregressive generation process. Given current observations and the task, the model produces (i) textual reasoning, (ii) low-level actions, and (iii) visual predictions of future frames. These imagined states, denoted by the fixed token \texttt{<Imagine:>}, are recursively fed back for internal reviewing and decision refinement. This mechanism allows iterative planning without environmental interaction.

\subsection{Training Pipeline}\label{sec:training}

\begin{figure}[t]
\centering
    \includegraphics[width=1\linewidth]{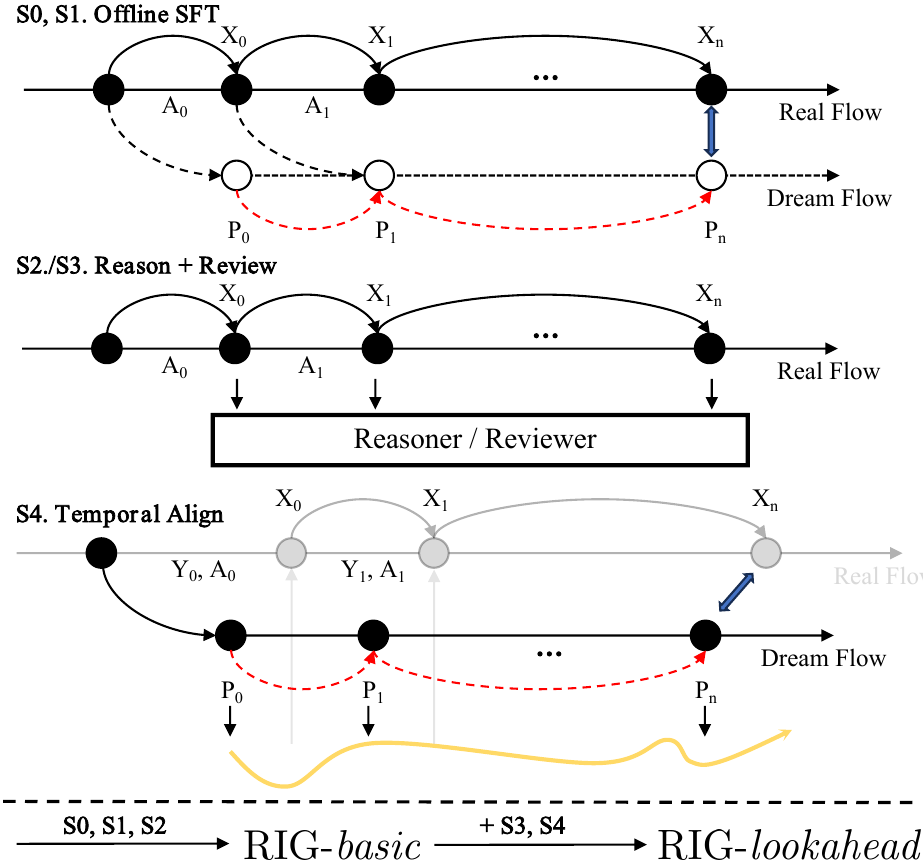}
    \caption{\textbf{Training pipeline of \methodname{}.} S0/S1 pretrain the model by aligning real and imagined flows. S2/S3 enhance reasoning and reviewing via GPT-4o relabeling. S4 aligns temporally predicted trajectories (dream flow) with environment-grounded traces.}
    \label{fig:training_pipeline}
\end{figure}

As illustrated in~\cref{fig:training_pipeline}, the training of \methodname{} proceeds in four progressive stages:

\begin{itemize}
    \item \textbf{S0/S1. Offline Supervised Fine-tuning (SFT):} The model learns to align the \textit{dream flow} (model-generated predictions) with the \textit{real flow} (observed data) through supervised learning. This phase improves visual state prediction quality, enhancing the accuracy of subsequent action decisions.  
    \newline\textit{Input:} past frame, past action, task.  
    \textit{Output:} subtask, next action, next frame.

    \item \textbf{S2. Reasoning Relabeling:} A two-step process enhances decision quality. (1) An environment-based evaluator filters high-quality trajectories. (2) GPT-4o acts as a \textbf{Reviewer} to generate explicit reasoning traces and refined labels.  
    \newline\textit{Input:} past frame, past action, task.  
    \textit{Output:} reasoning, next action, optionally lookahead reasoning, next frame.
    \item \textbf{S3. Review Relabeling:} The trained model interacts in the environment, and an evaluator filters poor trajectories. GPT-4o as a \textbf{Reviewer} analyzes the imagined traces and relabels corrections for better trajectory quality.  
    \newline\textit{Input:} past frame, past action, task, imagined frame (\texttt{<Imagine:>}).  
    \textit{Output:} lookahead reasoning
    , corrected action, next frame.

    \item \textbf{S4. Temporal Alignment:} An imagined dream trajectory is generated via the autoregressive model. The entire sequence is behavior-cloned into the real environment, enabling frame-by-frame alignment and relabeling via the Reasoner.  
    \newline\textit{Input:} dream trace (states/actions).  
    \textit{Output:} real visual alignment, updated reasoning annotations.
\end{itemize}

Stages 0–2 are used to train \methodname{}-\textit{basic}, while Stages 3–4 further enhance \methodname{}-\textit{lookahead} with imagination-based alignment and long-horizon correction.

\begin{figure}[h]
\centering
\includegraphics[width=1\linewidth]{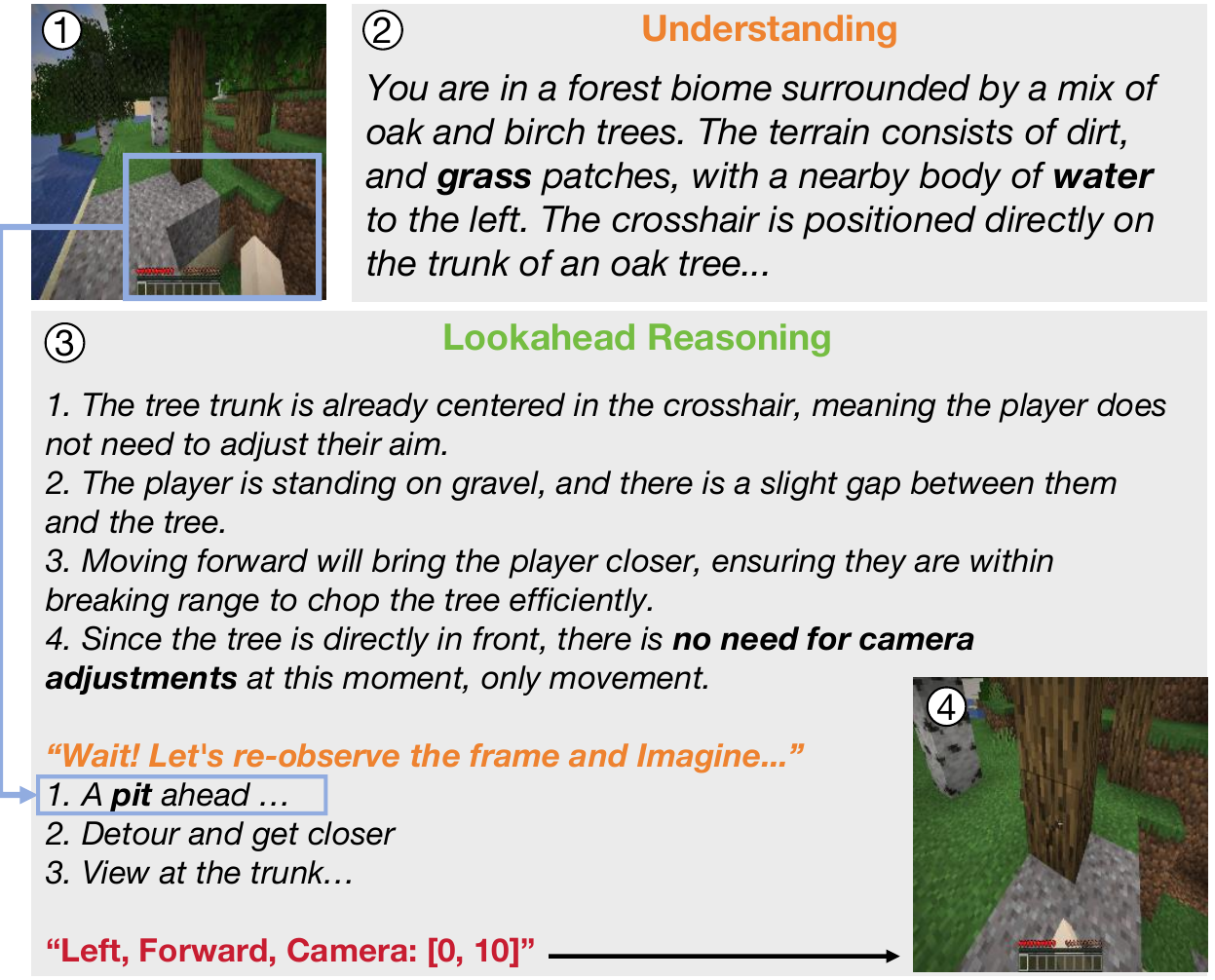}
\caption{\textbf{Qualitative example of lookahead and review.} The agent understands the environment (1–2), simulates future states (3), and refines its decision through internal review before acting (4), successfully avoiding a hidden hazard.}
\label{fig:example}
\end{figure}

\begin{table*}[ht]
\centering
\resizebox{\textwidth}{!}{
\begin{tabular}{lccccc}
\toprule
Method & Vision Encoder & Parameters & Vision Quality (Gen.) & MM Quality (Und.) & Evaluations \\
\midrule
\multicolumn{6}{l}{\textbf{Autoregressive (AR)}} \\
Emu3~\cite{wang2024emu3} & VQ (D) & 8B & 0.68 & -0.1 & POPE, SEEDBench-Img, VQAv2 (85.2, 68.2, 75.1) \\
LlamaGen~\cite{sun2024autoregressive} & VQ (D) & 111M, 343M, 775M, 1.4B, 3B & 0.68 & -0.34 & - \\
Chameleon~\cite{lu2023chameleon} & VQ (D) & 7B, 34B & 0.68 & -0.29 & VQAv2 (69.6) \\
Anole~\cite{chern2024anole} & VQ (D) & 7B & - & - & - \\
Janus~\cite{chen2025janus} & VQ (D) & 1.3B & 0.68 & -0.07 & POPE, VQAv2 (87, 77.3) \\
\midrule
\multicolumn{6}{l}{\textbf{AR + Diffusion}} \\
Show-o~\cite{xie2024show} & Magvitv2 (D/C), Clip-ViT (C) & 1.3B & 0.68 & -0.15 & POPE, VQAv2 (84.5, 74.7) \\
Transfusion~\cite{zhou2024transfusion} & VAE (C) & 0.16B, 0.37B, 0.76B, 1.4B, 7B & 0.68 & -0.01 & - \\
Fluid~\cite{fan2024fluid} & VQ (D), VAE (C) & 369M, 665M, 1.1B, 3.1B, 10.5B & 0.68 & 0.02 & - \\
\bottomrule
\end{tabular}
}
\caption{\textbf{Comparison of various unified multimodal methods}, categorized by their training approach (Autoregressive and AR + Diffusion), detailing vision encoder type, parameter scale, vision generation quality (GenEval SD3 8B), multimodal understanding quality, and evaluation benchmarks.}
\label{tab:method_comparison}
\end{table*}
\begin{table*}[t!]
    \centering
    \resizebox{\textwidth}{!}{
    \begin{tabular}{l|cc|cccc|ccc|ccc|ccc}
    \toprule
    \multirow{2}{*}{Tokenizer} & \multirow{2}{*}{Compression Ratio} & \multirow{2}{*}{Quantization} & \multicolumn{4}{c|}{MS-COCO} & \multicolumn{3}{c|}{ImageNet-1K} & \multicolumn{3}{c|}{FFHQ} & \multicolumn{3}{c}{CelebA-HQ} \\
    \cmidrule{4-16}
    & & & PSNR$\uparrow$ & SSIM$\uparrow$ & rFID$\downarrow$ & PSNR (Minecraft)$\uparrow$ & PSNR$\uparrow$ & SSIM$\uparrow$ & rFID$\downarrow$ & PSNR$\uparrow$ & SSIM$\uparrow$ & rFID$\downarrow$ & PSNR$\uparrow$ & SSIM$\uparrow$ & rFID$\downarrow$ \\
    \midrule
    Open-MAGVIT2~\cite{luo2024open} & $16\times16$ & LFQ & 30.06 & 0.502 & 6.649 & 27.21 & 29.62 & 0.398 & 2.701 & 31.77 & 0.774 & 1.994 & 32.36 & 0.844 & 2.865 \\
    LlamaGen~\cite{sun2024autoregressive} & $8\times8$ & VQ & 30.71 & 0.616 & 4.123 & 28.93 & 30.28 & 0.498 & 1.403 & 33.39 & 0.868 & 0.701 & 34.82 & 0.937 & 0.502 \\
    LlamaGen~\cite{sun2024autoregressive} & $16\times16$ & VQ & 29.93 & 0.491 & 6.077 & 27.06 & 29.81 & 0.448 & 1.657 & 31.58 & 0.772 & 1.366 & 32.18 & 0.837 & 1.113 \\
    Cosmos-Tokenizer-DI~\cite{agarwal2025cosmos} & $8\times8$ & FSQ & 31.74 & 0.730 & 4.564 & 30.84 & 31.73 & 0.725 & 1.841 & 35.35 & 0.892 & 0.555 & 37.77 & 0.948 & 0.261 \\
    Cosmos-Tokenizer-DI~\cite{agarwal2025cosmos} & $16\times16$ & FSQ & 30.74 & 0.591 & 12.252 & 29.91 & 30.69 & 0.582 & 6.529 & 33.17 & 0.808 & 7.663 & 33.86 & 0.854 & 5.953 \\
    Emu-3~\cite{wang2024emu3} & $16\times16$ & VQ & - & - & - & 24.16 & - & - & - & - & - & - & - & - & - \\
    \bottomrule
    \end{tabular}}
    \caption{\textbf{Comparison of Tokenizers across different benchmarks.} PSNR, SSIM, and rFID are measured on MS-COCO, ImageNet-1K, FFHQ, and CelebA-HQ datasets. PSNR for Minecraft images is provided separately.}
    \label{tab:tokenizer_comparison}
\end{table*}
\begin{table*}[!ht]
\vskip 0.1in
\centering
\resizebox{\textwidth}{!}{
    \begin{tabular}{p{2.4cm}|p{2.2cm}|p{2.2cm}|p{2.2cm}|p{2.2cm}|p{2.2cm}|p{2.2cm}|p{2.2cm}|p{2.2cm}}
    \toprule
    \textbf{Method} & VPT~\cite{openai2022vpt} & DreamerV3~\cite{hafner2023mastering} & DECKARD~\cite{nottingham2023embodied} & DEPS~\cite{wang2023describe} & Plan4MC~\cite{yuan2023plan4mc} & Voyager~\cite{wang2023voyager} &
    STEVE~\cite{lifshitz2023steve} & \textbf{RIG (Ours)}\\
    \midrule
    \textbf{Demos} 
      & Videos
      & None
      & Videos
      & None
      & None
      & None
      & Videos
      & Videos \\
    \midrule
    \textbf{Rewards}
      & Sparse
      & Dense
      & Sparse
      & None
      & Dense
      & None
      & None
      & None \\
    \midrule
    \textbf{Observations}
      & Pixels Only
      & \parbox[t]{2cm}{Pixels \\ \& Meta}
      & \parbox[t]{2cm}{Pixels \\ \& Inventory}
      & \parbox[t]{2cm}{Feedback \\ \& Inventory}
      & \parbox[t]{2cm}{Pixels \\ \& Meta}
      & \parbox[t]{2cm}{Feedback \\ \& Meta \\ \& Inventory}
      & \parbox[t]{2.2cm}{Pixels \\ \& Feedback \\ \& Meta \\ \& Inventory}
      & Pixels Only \\
    \midrule
    \textbf{Actions}
      & Keyboard \& Mouse
      & Discrete
      & Keyboard \& Mouse
      & Keyboard \& Mouse
      & Discrete
      & Code
      & Code
      & Keyboard \& Mouse \\
    \midrule
    \textbf{Reasoning}
      & 
      &
      &
      & \checkmark
      & 
      & \checkmark
      & \checkmark
      & \checkmark \\
    \midrule
    \textbf{Generation}
      & 
      &
      &
      & 
      & 
      & 
      & 
      & \checkmark \\
    \midrule
    \textbf{Extra Database}
      &
      &
      &
      &
      &
      \parbox[t]{2cm}{9}
      & \parbox[t]{2cm}{172}
      & \parbox[t]{2cm}{210}
      & - \\
    \bottomrule
    \end{tabular}
}
\caption{\textbf{Comparison between \textbf{RIG (Ours)}, and existing works.} This system-level comparison of LLM-based and RL-based methods focuses on data sources, reward setup, observation type, action representation, iterative planning, and skill database usage.}
\label{table:comparison_component}
\end{table*}
\begin{figure}[h]
\centering
\includegraphics[width=1\linewidth]{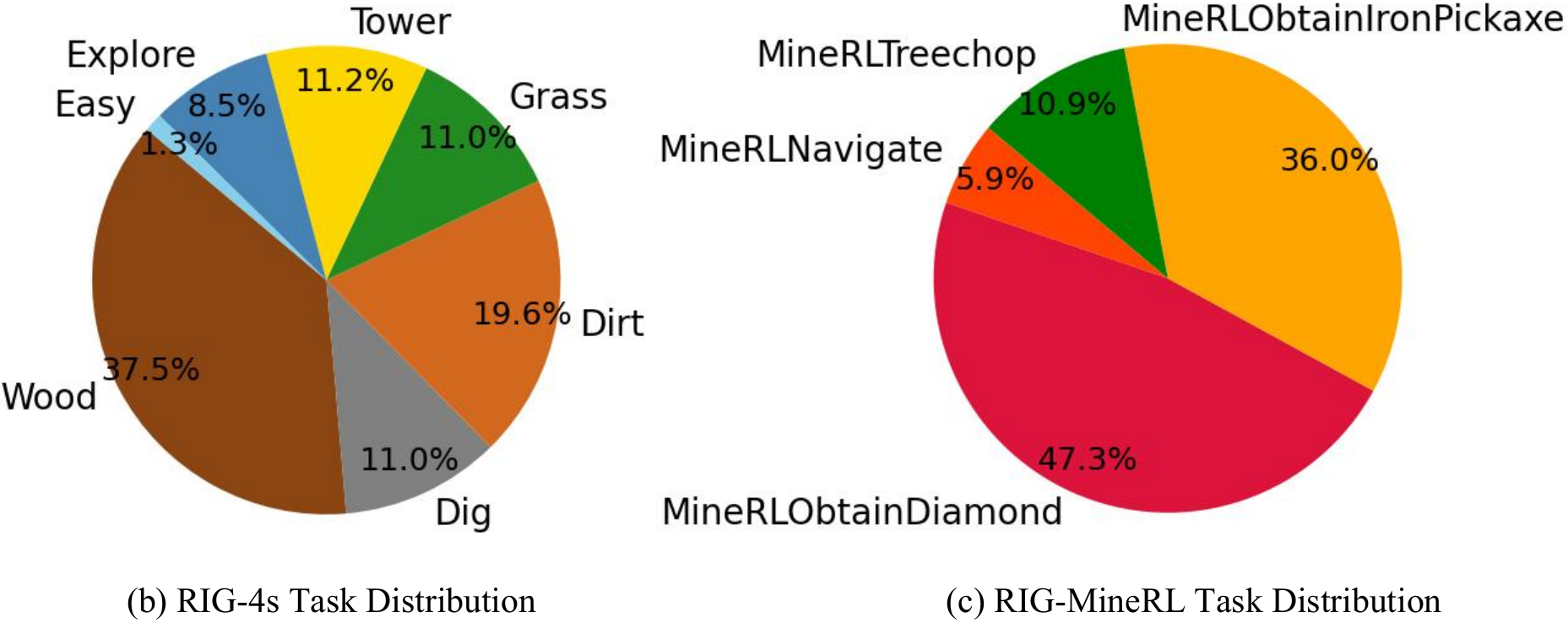}
\caption{\textbf{Task distribution.} Our datasets include various embodied tasks with varying complexity, ensuring strong generalization across downstream goals.}
\label{fig:data_distribution}
\end{figure}

\subsection{Data Distribution}\label{sec:data_distribution}

\cref{fig:data_distribution} visualizes the task distribution across our training datasets, which cover a spectrum of embodied scenarios such as resource collection, tower building, and exploration. As task complexity increases, we progressively expand the dataset size to ensure adequate supervision. Notably, harder tasks like building structures require significantly more data than simpler ones like gathering materials, highlighting the varying difficulty levels and skill composition in our training corpus.

\subsection{Component Comparison}\label{sec:component}
As summarized in \cref{tab:method_comparison} and \cref{table:comparison_component}, we compare \methodname{} to prior works along multiple dimensions, including input modality, action granularity, and reasoning capabilities. Unlike prior methods relying on handcrafted API actions or curated codebooks, \methodname{} operates solely on raw pixels and outputs keyboard-mouse controls, offering higher flexibility and lower task bias.
Notably, our design unifies reasoning and generation into a single transformer policy with self-review and imagination steps, offering better trajectory-level coherence and enabling multi-turn lookahead.

\subsection{Tokenizer and Base Model Selection}\label{sec:tokenizer}
We adopt LlamaGen 16$\times$16 VQ tokenizer and Janus-1.4B as our vision and language backbone. \cref{tab:tokenizer_comparison} reports their favorable reconstruction quality (PSNR 27.06) and semantic alignment. Janus uses a dual loss combining RGB and SigLIP-guided feature reconstruction, while LlamaGen provides discrete, compression-friendly tokens. Together, they form a scalable pipeline for visual imagination and reasoning, trained with simple cross-entropy objectives.

\subsection{Qualitative Results and Case Study}\label{sec:demo}

\cref{fig:example} demonstrates the full inference cycle of \methodname{}, where the agent understands the scene, reasons about its next move, simulates imagined outcomes, and conducts self-review before taking real action. In this wood-chopping task, the agent first identifies a tree in front, then reasons that moving forward seems viable. However, by simulating future states, it spots a hidden pit and triggers a self-correction: “\texttt{Wait! Let’s re-observe…}”. It updates its decision to \texttt{Left, Forward, Camera: [0, 10]: right}, successfully avoiding the hazard. This highlights our agent’s ability to perform proactive planning, visual forecasting, and risk-aware correction through imagination and reviewing.

\begin{figure*}[h]
\centering
\includegraphics[width=1\linewidth]{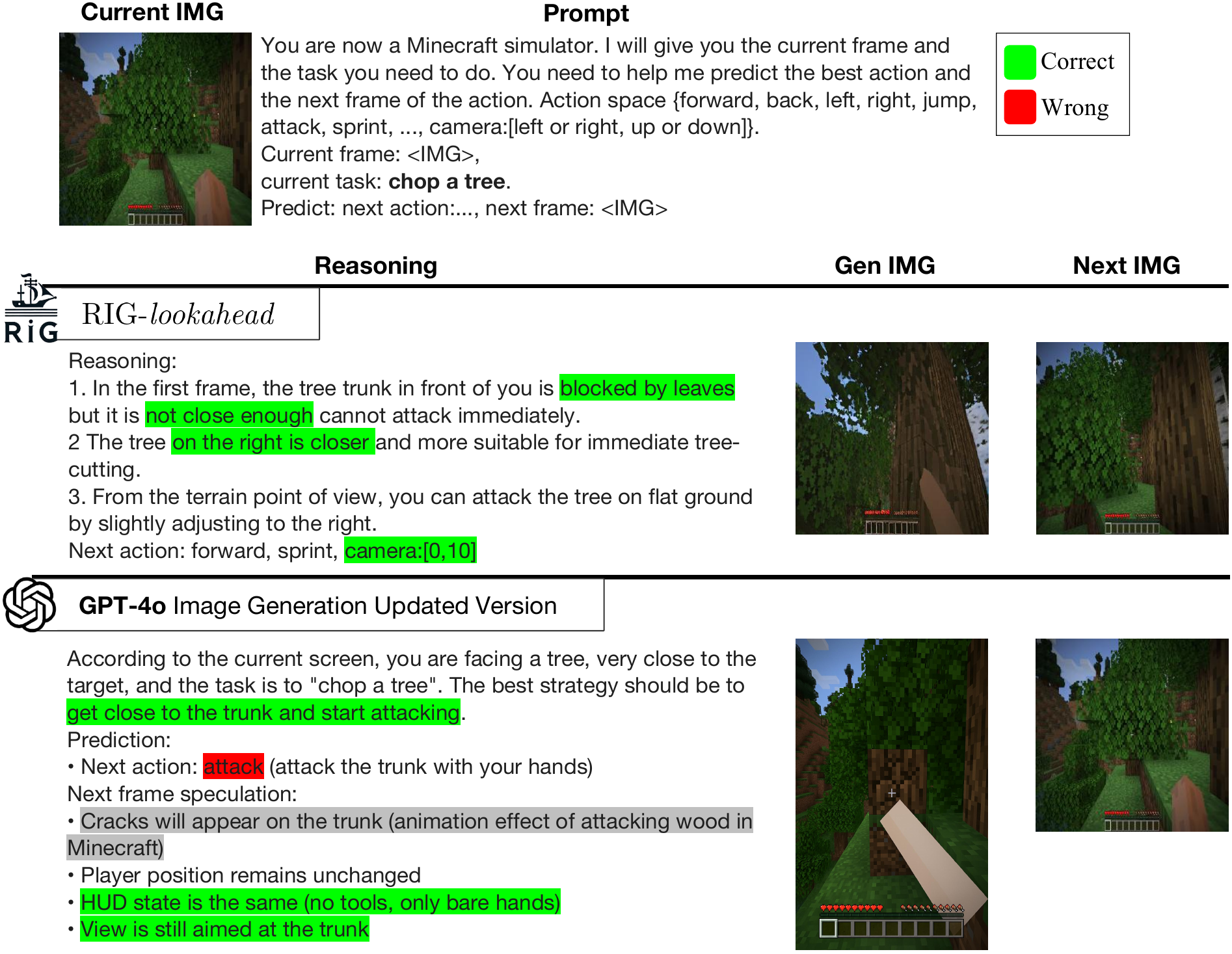}
\caption{\textbf{Case study comparison with GPT-4o.} Given the same input and prompt (\texttt{chop a tree}), \methodname{} reasons and imagines future states to choose a reachable tree and adjust position before acting. GPT-4o, despite high visual quality, misjudges the distance, executes an invalid action, and fails to revise its plan.}
\label{fig:case}
\end{figure*}

\cref{fig:case} further compares \methodname{}-\textit{lookahead} with GPT-4o image generation updated version. Both receive similar prompt and visual input. While GPT-4o generates a visually plausible prediction, it incorrectly judges the distance to the tree, prematurely issuing an \texttt{attack} command that leads to a deadlock. It continues to hallucinate progress without correcting the faulty assumption. In contrast, \methodname{} accurately detects that the tree is blocked and unreachable, reasons about terrain features, and adjusts its position before action. The generated image aligns with the actual environment response, showing stronger spatial consistency and robustness in long-horizon decision-making.

\subsection{Multi-Modal Understanding Evaluation}\label{sec:under_eval}
We further evaluate \methodname{} on the STEVE-21K~\cite{zheng2023steve} benchmark, testing its general world knowledge and Minecraft-specific understanding. Drawing from the Minecraft Wiki and Reddit corpus, the dataset spans multiple knowledge dimensions:

\begin{itemize}
    \item \textbf{World Understanding:} Terrain, entities, and biome behaviors.
    \item \textbf{Player Mechanics:} Combat, mobility, and health systems.
    \item \textbf{Survival Strategies:} Food sourcing, shelter, and threat avoidance.
    \item \textbf{Resource Management:} Gathering, mining, and inventory use.
    \item \textbf{Crafting and Construction:} Recipes and structural planning.
    \item \textbf{Tool Usage:} Equipment selection and upgrades.
\end{itemize}

We evaluate with 1000 QA pairs, categorized as: World \& Entities (332), Mechanics \& Survival (152), Knowledge \& Discovery (108), Crafting (219), Tools (169), and Miscellaneous (20). Our model demonstrates strong accuracy and reasoning coherence across categories.

\subsection{Multi-turn Visual Reasoning Format}~\label{sec:prompt_format}
To supervise step-level visual reasoning, we define a structured multi-turn dialogue format, as shown in below. Each entry logs the task instruction, prior action, current frame, reasoning, next action, and imagined future frame. This design aligns with autoregressive generation and supports fine-grained analysis and supervision.

\begin{itemize}
    \item \textbf{Task Instruction:} Natural language goal (\eg, ``build a tower'').
    \item \textbf{Previous Action:} Last executed action (\eg, ``camera:[0,10]'').
    \item \textbf{Current Frame:} Visual observation from the environment.
    \item \textbf{Step Reasoning:} Textual reasoning for the next decision.
    \item \textbf{Next Action:} Predicted action.
    \item \textbf{Next Frame:} Imagined visual result of the action.
\end{itemize}

\subsection{Environment Details}~\label{sec:env_detail}
\label{sup:Environment}

We use Minecraft as the testbed for embodied agents due to its open-ended nature and support for low-level human-like interactions. Agents act through egocentric RGB images and execute actions using keyboard and mouse inputs, making the environment ideal for sequential decision-making.

Our experiments are based on MineRL~\cite{guss2019minerl} v1.0 (Minecraft 1.16.5), which provides agents with first-person RGB observations and removes access to any privileged information. This version aligns with setups in prior works such as VPT~\cite{baker2022video} and STEVE-1~\cite{lifshitz2023steve}. Agents only perceive visual inputs and interact through low-level actions, resembling human play.

\subsection{Observation and Action Space}
\label{supsub:ObsAct}

The agent receives $640{\times}360$ RGB images rendered from a first-person view with a 70-degree field of view. When the inventory is opened, the GUI and mouse cursor are visible. No voxel, depth, or structured APIs (\eg, ``craft'', ``smelt'') are used.

As shown in \cref{table:comparison_component}, the action space includes 14 types of keyboard and mouse operations, covering movement (WASD), item use, inventory management, and camera rotation (yaw, pitch). These mirror human inputs and do not rely on high-level abstractions.

For the \texttt{camera} action, which is originally a 2D continuous vector $[a, b]$ representing pitch and yaw in $[-180^\circ, 180^\circ]$, we quantize it into discrete steps of 5 degrees to adapt to language models, where $a{<}0$/$a{>}0$ denote looking up/down and $b{<}0$/$b{>}0$ denote turning left/right; all other actions are binary (0/1), indicating whether to execute them.

\subsection{Environment Settings and Rules}
\label{supsub:Rules}

To ensure diversity and generalization, each episode is initialized with a random agent position and world seed, exposing the model to varying terrains, structures, and objectives.

We adapt the MineRL~\cite{guss2019minerl} environment to support instruction-based initialization via a custom \texttt{chat} action, enabling programmatic task injection and faster evaluation cycles.

\noindent\textbf{Programmatic Evaluation.} 
For instruction-following tasks, we fix the seed and biome across different models to ensure a fair comparison. Each experiment runs under conditions most favorable for the given instruction, and multiple seeds are used across different trials.

\begin{figure*}[t]
\begin{tcolorbox}[colback=white, colframe=black, boxrule=0.5pt, sharp corners=all, title=\textbf{Prompt Format for Minecraft Conversation Reasoning}]
\setlength{\parskip}{1ex}

\textbf{SYSTEM:}

You are assisting in generating step-by-step reasoning data for Minecraft embodied agents in complex tasks like \textbf{\textcolor{blue}{``build a tower''}}. Each step should contain (1) the reasoning process, (2) the predicted next action, and (3) the generated next image frame. We follow a multi-turn conversation structure with visual context.

\textbf{Input Format:}
\begin{itemize}
    \item \textbf{Task}: \textcolor{blue}{Given current task like ``build a tower''}
    \item \textbf{Previous Action}: \textcolor{orange}{e.g., camera:[0,10]}
    \item \textbf{Current Frame}: Image observation
\end{itemize}

\textbf{Human Prompt (Input):}
\begin{verbatim}
<image>
Please make reasoning of action and predict the next action based on the 
current frame, current task and previous action, and generate the next frame 
based on the current frame and the next action.
The current task: build a tower. 
The previous action: camera:[0,10].
\end{verbatim}

\textbf{RIG Output (Response):}
\begin{tcolorbox}[colback=gray!5, colframe=gray!50!black]
\textbf{The reasoning:} In the current frame, the task is to \textcolor{blue}{build a tower}, but no suitable material is directly targeted. The player may be exploring for logs or stone. The previous action moved the camera upward to scout surroundings...

\textbf{Next action:} \textcolor{red}{left, sprint}, and next frame: \texttt{<image>}
\end{tcolorbox}

\textbf{Special Cases: Lookahead and Imagination}
\begin{itemize}
    \item When the model performs internal reflection or imagines the future, use the special token: \texttt{\textcolor{purple}{<Imagine:>}} to prefix the reasoning block.
    \item Example: \texttt{\textcolor{purple}{<Imagine:>}} Reasoning forward $n$ steps into future environment outcomes.
\end{itemize}

\textbf{Fields to be included in JSON:}
\begin{itemize}
    \item \texttt{name}, \texttt{id}, \texttt{action}, \texttt{images}, \texttt{conversations}, \texttt{subtask} (optional)
\end{itemize}

\textbf{Example JSON Structure:}
\begin{verbatim}
{
  "name": "build_a_tower_seed203",
  "id": 10,
  "action": "left, sprint",
  "conversations": [
    {
      "from": "human",
      "value": "<image>\n <Imagine:> Please make reasoning of action... 
      task: build a tower...",
      "images": ["..._10_current.png"]
    },
    {
      "from": "RIG",
      "value": "The reasoning: ... Next action: left, sprint, 
      sand next frame: <image>",
      "images": ["..._10_next.png"]
    }
  ]
}
\end{verbatim}

\end{tcolorbox}
\label{fig:prompt}
\end{figure*}

\noindent\textbf{Long-Horizon Tasks with Command Switching.} In these scenarios, agents operate in a shared environment initialized with a common seed and biome. To reduce noise and stabilize training/testing, we apply the following environment rules:
\begin{itemize}
    \small
    \item \texttt{/difficulty peaceful}: Disables hostile mobs.
    \item \texttt{/gamerule doDaylightCycle false}: Locks the environment in perpetual daytime.
    \item \texttt{/gamerule keepInventory true}: Prevents item loss upon death.
\end{itemize}

These configurations help maintain consistency across runs while preserving the task’s difficulty and realism. They also support efficient benchmarking of embodied reasoning and planning in long-horizon, open-ended environments. 

\clearpage

{
    \small
    \bibliographystyle{ieeenat_fullname}
    \bibliography{main}

\begin{thebibliography}{48}
\providecommand{\natexlab}[1]{#1}
\providecommand{\url}[1]{\texttt{#1}}
\expandafter\ifx\csname urlstyle\endcsname\relax
  \providecommand{\doi}[1]{doi: #1}\else
  \providecommand{\doi}{doi: \begingroup \urlstyle{rm}\Url}\fi

\bibitem[Agarwal et~al.(2025)Agarwal, Ali, Bala, Balaji, Barker, Cai, Chattopadhyay, Chen, Cui, Ding, et~al.]{agarwal2025cosmos}
Niket Agarwal, Arslan Ali, Maciej Bala, Yogesh Balaji, Erik Barker, Tiffany Cai, Prithvijit Chattopadhyay, Yongxin Chen, Yin Cui, Yifan Ding, et~al.
\newblock Cosmos world foundation model platform for physical ai.
\newblock \emph{arXiv preprint arXiv:2501.03575}, 2025.

\bibitem[Baker et~al.(2022{\natexlab{a}})Baker, Akkaya, Zhokhov, Huizinga, Tang, Ecoffet, Houghton, Sampedro, and Clune]{openai2022vpt}
Bowen Baker, Ilge Akkaya, Peter Zhokhov, Joost Huizinga, Jie Tang, Adrien Ecoffet, Brandon Houghton, Raul Sampedro, and Jeff Clune.
\newblock Video pretraining (vpt): Learning to act by watching unlabeled online videos.
\newblock \emph{arXiv preprint arXiv: Arxiv-2206.11795}, 2022{\natexlab{a}}.

\bibitem[Baker et~al.(2022{\natexlab{b}})Baker, Akkaya, Zhokov, Huizinga, Tang, Ecoffet, Houghton, Sampedro, and Clune]{baker2022video}
Bowen Baker, Ilge Akkaya, Peter Zhokov, Joost Huizinga, Jie Tang, Adrien Ecoffet, Brandon Houghton, Raul Sampedro, and Jeff Clune.
\newblock Video pretraining (vpt): Learning to act by watching unlabeled online videos.
\newblock \emph{Advances in Neural Information Processing Systems}, 35:\penalty0 24639--24654, 2022{\natexlab{b}}.

\bibitem[Brohan et~al.(2023)Brohan, Brown, Carbajal, Chebotar, Dabis, Finn, Gopalakrishnan, Hausman, Herzog, Hsu, et~al.]{brohan2022rt}
Anthony Brohan, Noah Brown, Justice Carbajal, Yevgen Chebotar, Joseph Dabis, Chelsea Finn, Keerthana Gopalakrishnan, Karol Hausman, Alex Herzog, Jasmine Hsu, et~al.
\newblock {RT-1}: Robotics transformer for real-world control at scale.
\newblock In \emph{RSS}, 2023.

\bibitem[Brown et~al.(2020)Brown, Mann, Ryder, Subbiah, Kaplan, Dhariwal, Neelakantan, Shyam, Sastry, Askell, et~al.]{brown2020language}
Tom Brown, Benjamin Mann, Nick Ryder, Melanie Subbiah, Jared~D Kaplan, Prafulla Dhariwal, Arvind Neelakantan, Pranav Shyam, Girish Sastry, Amanda Askell, et~al.
\newblock Language models are few-shot learners.
\newblock \emph{Advances in neural information processing systems}, 33:\penalty0 1877--1901, 2020.

\bibitem[Cai et~al.(2023{\natexlab{a}})Cai, Wang, Ma, Liu, and Liang]{cai2023open}
Shaofei Cai, Zihao Wang, Xiaojian Ma, Anji Liu, and Yitao Liang.
\newblock Open-world multi-task control through goal-aware representation learning and adaptive horizon prediction.
\newblock In \emph{Proceedings of the IEEE/CVF Conference on Computer Vision and Pattern Recognition}, pages 13734--13744, 2023{\natexlab{a}}.

\bibitem[Cai et~al.(2023{\natexlab{b}})Cai, Zhang, Wang, Ma, Liu, and Liang]{cai2023groot}
Shaofei Cai, Bowei Zhang, Zihao Wang, Xiaojian Ma, Anji Liu, and Yitao Liang.
\newblock Groot: Learning to follow instructions by watching gameplay videos.
\newblock \emph{arXiv preprint arXiv:2310.08235}, 2023{\natexlab{b}}.

\bibitem[Chen et~al.(2025)Chen, Wu, Liu, Pan, Liu, Xie, Yu, and Ruan]{chen2025janus}
Xiaokang Chen, Zhiyu Wu, Xingchao Liu, Zizheng Pan, Wen Liu, Zhenda Xie, Xingkai Yu, and Chong Ruan.
\newblock Janus-pro: Unified multimodal understanding and generation with data and model scaling.
\newblock \emph{arXiv preprint arXiv:2501.17811}, 2025.

\bibitem[Chern et~al.(2024)Chern, Su, Ma, and Liu]{chern2024anole}
Ethan Chern, Jiadi Su, Yan Ma, and Pengfei Liu.
\newblock Anole: An open, autoregressive, native large multimodal models for interleaved image-text generation.
\newblock \emph{arXiv preprint arXiv:2407.06135}, 2024.

\bibitem[Contributors(2023)]{2023xtuner}
XTuner Contributors.
\newblock Xtuner: A toolkit for efficiently fine-tuning llm.
\newblock \url{https://github.com/InternLM/xtuner}, 2023.

\bibitem[Fan et~al.(2022)Fan, Wang, Jiang, Mandlekar, Yang, Zhu, Tang, Huang, Zhu, and Anandkumar]{fan2022minedojo}
Linxi Fan, Guanzhi Wang, Yunfan Jiang, Ajay Mandlekar, Yuncong Yang, Haoyi Zhu, Andrew Tang, De-An Huang, Yuke Zhu, and Anima Anandkumar.
\newblock Minedojo: Building open-ended embodied agents with internet-scale knowledge.
\newblock \emph{Advances in Neural Information Processing Systems}, 35:\penalty0 18343--18362, 2022.

\bibitem[Fan et~al.(2024)Fan, Li, Qin, Li, Sun, Rubinstein, Sun, He, and Tian]{fan2024fluid}
Lijie Fan, Tianhong Li, Siyang Qin, Yuanzhen Li, Chen Sun, Michael Rubinstein, Deqing Sun, Kaiming He, and Yonglong Tian.
\newblock Fluid: Scaling autoregressive text-to-image generative models with continuous tokens.
\newblock \emph{arXiv preprint arXiv:2410.13863}, 2024.

\bibitem[Guss et~al.(2019)Guss, Houghton, Topin, Wang, Codel, Veloso, and Salakhutdinov]{guss2019minerl}
William~H Guss, Brandon Houghton, Nicholay Topin, Phillip Wang, Cayden Codel, Manuela Veloso, and Ruslan Salakhutdinov.
\newblock Minerl: a large-scale dataset of minecraft demonstrations.
\newblock In \emph{Proceedings of the 28th International Joint Conference on Artificial Intelligence}, pages 2442--2448, 2019.

\bibitem[Hafner et~al.(2020)Hafner, Lillicrap, Ba, and Norouzi]{hafner2019dream}
Danijar Hafner, Timothy Lillicrap, Jimmy Ba, and Mohammad Norouzi.
\newblock Dream to control: Learning behaviors by latent imagination.
\newblock In \emph{ICLR}, 2020.

\bibitem[Hafner et~al.(2021)Hafner, Lillicrap, Norouzi, and Ba]{hafner2020mastering}
Danijar Hafner, Timothy Lillicrap, Mohammad Norouzi, and Jimmy Ba.
\newblock Mastering atari with discrete world models.
\newblock In \emph{ICLR}, 2021.

\bibitem[Hafner et~al.(2023)Hafner, Pasukonis, Ba, and Lillicrap]{hafner2023mastering}
Danijar Hafner, Jurgis Pasukonis, Jimmy Ba, and Timothy Lillicrap.
\newblock Mastering diverse domains through world models.
\newblock \emph{arXiv preprint arXiv:2301.04104}, 2023.

\bibitem[Hansen et~al.(2022)Hansen, Wang, and Su]{hansen2022temporal}
Nicklas Hansen, Xiaolong Wang, and Hao Su.
\newblock Temporal difference learning for model predictive control.
\newblock In \emph{ICML}, 2022.

\bibitem[Johnson et~al.(2016)Johnson, Hofmann, Hutton, and Bignell]{johnson2016malmo}
Matthew Johnson, Katja Hofmann, Tim Hutton, and David Bignell.
\newblock The malmo platform for artificial intelligence experimentation.
\newblock In \emph{Ijcai}, pages 4246--4247, 2016.

\bibitem[Kaiser et~al.(2020)Kaiser, Babaeizadeh, Milos, Osinski, Campbell, Czechowski, Erhan, Finn, Kozakowski, Levine, et~al.]{kaiser2019model}
Lukasz Kaiser, Mohammad Babaeizadeh, Piotr Milos, Blazej Osinski, Roy~H Campbell, Konrad Czechowski, Dumitru Erhan, Chelsea Finn, Piotr Kozakowski, Sergey Levine, et~al.
\newblock Model-based reinforcement learning for atari.
\newblock In \emph{ICLR}, 2020.

\bibitem[Lifshitz et~al.(2023)Lifshitz, Paster, Chan, Ba, and McIlraith]{lifshitz2023steve}
Shalev Lifshitz, Keiran Paster, Harris Chan, Jimmy Ba, and Sheila McIlraith.
\newblock Steve-1: A generative model for text-to-behavior in minecraft.
\newblock \emph{arXiv preprint arXiv:2306.00937}, 2023.

\bibitem[Lin et~al.(2023)Lin, Du, Watkins, Hafner, Abbeel, Klein, and Dragan]{lin2023learning}
Jessy Lin, Yuqing Du, Olivia Watkins, Danijar Hafner, Pieter Abbeel, Dan Klein, and Anca Dragan.
\newblock Learning to model the world with language.
\newblock 2023.

\bibitem[Lu et~al.(2023)Lu, Peng, Cheng, Galley, Chang, Wu, Zhu, and Gao]{lu2023chameleon}
Pan Lu, Baolin Peng, Hao Cheng, Michel Galley, Kai-Wei Chang, Ying~Nian Wu, Song-Chun Zhu, and Jianfeng Gao.
\newblock Chameleon: Plug-and-play compositional reasoning with large language models.
\newblock \emph{arXiv preprint arXiv:2304.09842}, 2023.

\bibitem[Luo et~al.(2024)Luo, Shi, Ge, Yang, Wang, and Shan]{luo2024open}
Zhuoyan Luo, Fengyuan Shi, Yixiao Ge, Yujiu Yang, Limin Wang, and Ying Shan.
\newblock Open-magvit2: An open-source project toward democratizing auto-regressive visual generation.
\newblock \emph{arXiv preprint arXiv:2409.04410}, 2024.

\bibitem[Mendonca et~al.(2023)Mendonca, Bahl, and Pathak]{mendonca2023structured}
Russell Mendonca, Shikhar Bahl, and Deepak Pathak.
\newblock Structured world models from human videos.
\newblock In \emph{RSS}, 2023.

\bibitem[Micheli et~al.(2023)Micheli, Alonso, and Fleuret]{micheli2022transformers}
Vincent Micheli, Eloi Alonso, and Fran{\c{c}}ois Fleuret.
\newblock Transformers are sample efficient world models.
\newblock In \emph{ICLR}, 2023.

\bibitem[Nottingham et~al.(2023)Nottingham, Ammanabrolu, Suhr, Choi, Hajishirzi, Singh, and Fox]{nottingham2023embodied}
Kolby Nottingham, Prithviraj Ammanabrolu, Alane Suhr, Yejin Choi, Hannaneh Hajishirzi, Sameer Singh, and Roy Fox.
\newblock Do embodied agents dream of pixelated sheep?: Embodied decision making using language guided world modelling.
\newblock \emph{arXiv preprint}, 2023.

\bibitem[Oh et~al.(2015)Oh, Guo, Lee, Lewis, and Singh]{oh2015actionconditional}
Junhyuk Oh, Xiaoxiao Guo, Honglak Lee, Richard Lewis, and Satinder Singh.
\newblock Action-conditional video prediction using deep networks in atari games.
\newblock In \emph{NeurIPS}, 2015.

\bibitem[OpenAI(2023)]{gpt4v}
OpenAI.
\newblock Gpt-4v(ision) system card.
\newblock 2023.

\bibitem[Reed et~al.(2022)Reed, Zolna, Parisotto, Colmenarejo, Novikov, Barth-Maron, Gimenez, Sulsky, Kay, Springenberg, et~al.]{reed2022generalist}
Scott Reed, Konrad Zolna, Emilio Parisotto, Sergio~Gomez Colmenarejo, Alexander Novikov, Gabriel Barth-Maron, Mai Gimenez, Yury Sulsky, Jackie Kay, Jost~Tobias Springenberg, et~al.
\newblock A generalist agent.
\newblock \emph{arXiv preprint arXiv:2205.06175}, 2022.

\bibitem[Schrittwieser et~al.(2020)Schrittwieser, Antonoglou, Hubert, Simonyan, Sifre, Schmitt, Guez, Lockhart, Hassabis, Graepel, et~al.]{schrittwieser2020mastering}
Julian Schrittwieser, Ioannis Antonoglou, Thomas Hubert, Karen Simonyan, Laurent Sifre, Simon Schmitt, Arthur Guez, Edward Lockhart, Demis Hassabis, Thore Graepel, et~al.
\newblock Mastering atari, go, chess and shogi by planning with a learned model.
\newblock \emph{Nature}, 588\penalty0 (7839):\penalty0 604--609, 2020.

\bibitem[Sun et~al.(2024)Sun, Jiang, Chen, Zhang, Peng, Luo, and Yuan]{sun2024autoregressive}
Peize Sun, Yi Jiang, Shoufa Chen, Shilong Zhang, Bingyue Peng, Ping Luo, and Zehuan Yuan.
\newblock Autoregressive model beats diffusion: Llama for scalable image generation.
\newblock \emph{arXiv preprint arXiv:2406.06525}, 2024.

\bibitem[Touvron et~al.(2023)Touvron, Martin, Stone, Albert, Almahairi, Babaei, Bashlykov, Batra, Bhargava, Bhosale, et~al.]{touvron2023llama}
Hugo Touvron, Louis Martin, Kevin Stone, Peter Albert, Amjad Almahairi, Yasmine Babaei, Nikolay Bashlykov, Soumya Batra, Prajjwal Bhargava, Shruti Bhosale, et~al.
\newblock Llama 2: Open foundation and fine-tuned chat models.
\newblock \emph{arXiv preprint arXiv:2307.09288}, 2023.

\bibitem[Wang et~al.(2023{\natexlab{a}})Wang, Xie, Jiang, Mandlekar, Xiao, Zhu, Fan, and Anandkumar]{wang2023voyager}
Guanzhi Wang, Yuqi Xie, Yunfan Jiang, Ajay Mandlekar, Chaowei Xiao, Yuke Zhu, Linxi Fan, and Anima Anandkumar.
\newblock Voyager: An open-ended embodied agent with large language models.
\newblock \emph{arXiv preprint arXiv:2305.16291}, 2023{\natexlab{a}}.

\bibitem[Wang et~al.(2024)Wang, Zhang, Luo, Sun, Cui, Wang, Zhang, Wang, Li, Yu, et~al.]{wang2024emu3}
Xinlong Wang, Xiaosong Zhang, Zhengxiong Luo, Quan Sun, Yufeng Cui, Jinsheng Wang, Fan Zhang, Yueze Wang, Zhen Li, Qiying Yu, et~al.
\newblock Emu3: Next-token prediction is all you need.
\newblock \emph{arXiv preprint arXiv:2409.18869}, 2024.

\bibitem[Wang et~al.(2023{\natexlab{b}})Wang, Cai, Liu, Jin, Hou, Zhang, Lin, He, Zheng, Yang, et~al.]{wang2023jarvis}
Zihao Wang, Shaofei Cai, Anji Liu, Yonggang Jin, Jinbing Hou, Bowei Zhang, Haowei Lin, Zhaofeng He, Zilong Zheng, Yaodong Yang, et~al.
\newblock Jarvis-1: Open-world multi-task agents with memory-augmented multimodal language models.
\newblock \emph{arXiv preprint arXiv:2311.05997}, 2023{\natexlab{b}}.

\bibitem[Wang et~al.(2023{\natexlab{c}})Wang, Cai, Liu, Ma, and Liang]{wang2023describe}
Zihao Wang, Shaofei Cai, Anji Liu, Xiaojian Ma, and Yitao Liang.
\newblock Describe, explain, plan and select: Interactive planning with large language models enables open-world multi-task agents.
\newblock \emph{arXiv preprint arXiv:2302.01560}, 2023{\natexlab{c}}.

\bibitem[Wu et~al.(2023)Wu, Ma, Deng, and Long]{wu2024pre}
Jialong Wu, Haoyu Ma, Chaoyi Deng, and Mingsheng Long.
\newblock Pre-training contextualized world models with in-the-wild videos for reinforcement learning.
\newblock In \emph{NeurIPS}, 2023.

\bibitem[Xie et~al.(2024)Xie, Mao, Bai, Zhang, Wang, Lin, Gu, Chen, Yang, and Shou]{xie2024show}
Jinheng Xie, Weijia Mao, Zechen Bai, David~Junhao Zhang, Weihao Wang, Kevin~Qinghong Lin, Yuchao Gu, Zhijie Chen, Zhenheng Yang, and Mike~Zheng Shou.
\newblock Show-o: One single transformer to unify multimodal understanding and generation.
\newblock \emph{arXiv preprint arXiv:2408.12528}, 2024.

\bibitem[Yu et~al.(2023)Yu, Shi, Pasunuru, Muller, Golovneva, Wang, Babu, Tang, Karrer, Sheynin, et~al.]{yu2023scaling}
Lili Yu, Bowen Shi, Ramakanth Pasunuru, Benjamin Muller, Olga Golovneva, Tianlu Wang, Arun Babu, Binh Tang, Brian Karrer, Shelly Sheynin, et~al.
\newblock Scaling autoregressive multi-modal models: Pretraining and instruction tuning.
\newblock \emph{arXiv preprint arXiv:2309.02591}, 2\penalty0 (3):\penalty0 3, 2023.

\bibitem[Yuan et~al.(2023)Yuan, Zhang, Wang, Xie, Cai, Dong, and Lu]{yuan2023plan4mc}
Haoqi Yuan, Chi Zhang, Hongcheng Wang, Feiyang Xie, Penglin Cai, Hao Dong, and Zongqing Lu.
\newblock Plan4mc: Skill reinforcement learning and planning for open-world minecraft tasks.
\newblock \emph{arXiv preprint arXiv:2303.16563}, 2023.

\bibitem[Zhai et~al.(2023)Zhai, Mustafa, Kolesnikov, and Beyer]{zhai2023sigmoid}
Xiaohua Zhai, Basil Mustafa, Alexander Kolesnikov, and Lucas Beyer.
\newblock Sigmoid loss for language image pre-training.
\newblock In \emph{Proceedings of the IEEE/CVF International Conference on Computer Vision}, pages 11975--11986, 2023.

\bibitem[Zhang et~al.(2023)Zhang, Cai, Fu, Yuan, and Lu]{zhang2023creative}
Chi Zhang, Penglin Cai, Yuhui Fu, Haoqi Yuan, and Zongqing Lu.
\newblock Creative agents: Empowering agents with imagination for creative tasks.
\newblock \emph{arXiv preprint arXiv:2312.02519}, 2023.

\bibitem[Zhao et~al.(2023)Zhao, Chai, Wang, Boyi, Hao, Cao, Ye, Hwang, and Wang]{zhao2023see}
Zhonghan Zhao, Wenhao Chai, Xuan Wang, Li Boyi, Shengyu Hao, Shidong Cao, Tian Ye, Jenq-Neng Hwang, and Gaoang Wang.
\newblock See and think: Embodied agent in virtual environment.
\newblock \emph{arXiv preprint arXiv:2311.15209}, 2023.

\bibitem[Zhao et~al.(2024{\natexlab{a}})Zhao, Chen, Guo, Chai, Ye, Zhang, and Wang]{zhao2024hierarchical}
Zhonghan Zhao, Kewei Chen, Dongxu Guo, Wenhao Chai, Tian Ye, Yanting Zhang, and Gaoang Wang.
\newblock Hierarchical auto-organizing system for open-ended multi-agent navigation.
\newblock \emph{arXiv preprint arXiv:2403.08282}, 2024{\natexlab{a}}.

\bibitem[Zhao et~al.(2024{\natexlab{b}})Zhao, Ma, Chai, Wang, Chen, Guo, Zhang, Wang, and Wang]{zhao2024we}
Zhonghan Zhao, Ke Ma, Wenhao Chai, Xuan Wang, Kewei Chen, Dongxu Guo, Yanting Zhang, Hongwei Wang, and Gaoang Wang.
\newblock Do we really need a complex agent system? distill embodied agent into a single model.
\newblock \emph{arXiv preprint arXiv:2404.04619}, 2024{\natexlab{b}}.

\bibitem[Zheng et~al.(2023)Zheng, Liu, Feng, and Lu]{zheng2023steve}
Sipeng Zheng, Jiazheng Liu, Yicheng Feng, and Zongqing Lu.
\newblock Steve-eye: Equipping llm-based embodied agents with visual perception in open worlds.
\newblock \emph{arXiv preprint arXiv:2310.13255}, 2023.

\bibitem[Zhou et~al.(2024{\natexlab{a}})Zhou, Yu, Babu, Tirumala, Yasunaga, Shamis, Kahn, Ma, Zettlemoyer, and Levy]{zhou2024transfusion}
Chunting Zhou, Lili Yu, Arun Babu, Kushal Tirumala, Michihiro Yasunaga, Leonid Shamis, Jacob Kahn, Xuezhe Ma, Luke Zettlemoyer, and Omer Levy.
\newblock Transfusion: Predict the next token and diffuse images with one multi-modal model.
\newblock \emph{arXiv preprint arXiv:2408.11039}, 2024{\natexlab{a}}.

\bibitem[Zhou et~al.(2024{\natexlab{b}})Zhou, Qin, Yin, Huang, Zhang, Sheng, Qiao, and Shao]{zhou2024minedreamer}
Enshen Zhou, Yiran Qin, Zhenfei Yin, Yuzhou Huang, Ruimao Zhang, Lu Sheng, Yu Qiao, and Jing Shao.
\newblock Minedreamer: Learning to follow instructions via chain-of-imagination for simulated-world control.
\newblock \emph{arXiv preprint arXiv:2403.12037}, 2024{\natexlab{b}}.

\end{thebibliography}
}
\end{document}